\documentclass[journal]{IEEEtran}
\usepackage{makecell}

\usepackage{tabularx,booktabs}

\usepackage{ifpdf}
 \ifpdf
 \else
 \fi

\usepackage{cite}

\ifCLASSINFOpdf
   \usepackage[pdftex]{graphicx}
\else
\fi

\usepackage{array}

\usepackage{algorithmic}
\usepackage{algorithm}

\usepackage{float}

\ifCLASSOPTIONcompsoc 
  \usepackage[caption=false,font=normalsize,labelfont=sf,textfont=sf]{subfig}
\else
  \usepackage[caption=false,font=footnotesize]{subfig}
\fi

\usepackage{url}

\makeatletter
\let\NAT@parse\undefined
\makeatother
\usepackage{hyperref}  %hyperref still needs to be put at the end!

\usepackage{times}
\usepackage{epsfig}
\usepackage{graphicx}
\usepackage{amssymb}
\usepackage{multirow}
\usepackage{color}

\usepackage{amsthm,amsmath,amssymb,amsfonts}
\usepackage{mathrsfs}

\ifCLASSINFOpdf
\else
\fi
\usepackage[utf8]{inputenc}
\usepackage{cleveref}
\crefname{section}{§}{§§}
\Crefname{section}{§}{§§}

\begin{document}
\title{G$^2$DA: Geometry-Guided Dual-Alignment Learning for RGB-Infrared Person Re-Identification}

\author{Lin Wan, Zongyuan Sun, Qianyan Jing, Yehansen Chen, Lijing Lu, and Zhihang Li % <-this % stops a space
\thanks{(Corresponding author: Zhihang Li)}%
\thanks{Lin Wan, Zongyuan Sun, Qianyan Jing and Yehansen Chen are with the School of Geography and Information Engineering, China University of Geoscience, Wuhan 430078, China (e-mail: wanlin@cug.edu.cn; sunzongyuan@cug.edu.cn; jingqianyan@cug.edu.cn; chenyehansen@cug.edu.cn).}% 
\thanks{Lijing Lu and Zhihang Li are with University of Chinese Academy of Sciences, Beijing 100190, China (e-mail: lulijing2019@ia.ac.cn; lizhihang.cas@gmail.com).}% <-this % stops a space
}


\markboth{~Vol.~X, No.~X, XXX~2021}%
{Shell \MakeLowercase{\textit{et al.}}: Bare Demo of IEEEtran.cls for IEEE Journals}



\maketitle
\begin{abstract}
RGB-Infrared (IR) person re-identification aims to retrieve person-of-interest from heterogeneous cameras, easily suffering from large image modality discrepancy caused by different sensing wavelength ranges. Existing work usually minimizes such discrepancy by aligning domain distribution of global features, while neglecting the intra-modality structural relations between semantic parts. This could result in the network overly focusing on local cues, without considering long-range body part dependencies, leading to meaningless region representations. In this paper, we propose a graph-enabled distribution matching solution, dubbed Geometry-Guided Dual-Alignment (G$^2$DA) learning, for RGB-IR ReID. It can jointly encourage the cross-modal consistency between part semantics and structural relations for fine-grained modality alignment by solving a graph matching task within a multi-scale skeleton graph that embeds human topology information. Specifically, we propose to build a semantic-aligned complete graph into which all cross-modality images can be mapped via a pose-adaptive graph construction mechanism. This graph represents extracted whole-part features by nodes and expresses the node-wise similarities with associated edges. To achieve the graph-based dual-alignment learning, an Optimal Transport (OT) based structured metric is further introduced to simultaneously measure point-wise relations and group-wise structural similarities across modalities. By minimizing the cost of a inter-modality transport plan, G$^2$DA can learn a consistent and discriminative feature subspace for cross-modality image retrieval. Furthermore, we advance a Message Fusion Attention (MFA) mechanism to adaptively reweight the information flow of semantic propagation, effectively strengthening the discriminability of extracted semantic features. Extensive experiments on two standard benchmark datasets validate the superiority of G$^2$DA, yielding competitive performance against previous state-of-the-arts.
\end{abstract}

\begin{IEEEkeywords}
Person re-identification, cross-modality matching, optimal transport, feature alignment, channel exchange.
\end{IEEEkeywords}

% \fi
\IEEEpeerreviewmaketitle

%----------- 1. Introduction ---------------
\section{Introduction}
\IEEEPARstart{P}{erson} re-identification (ReID) refers to a large-scale image retrieval task that aims to match specific pedestrians from multiple non-overlapping camera views \cite{Ye2021VisibleInfraredPR,Wang2019RGBInfraredCP}. It has picked up momentum in recent years for its widespread use in video surveillance and public security \cite{Khan2016PersonRF,article}. Currently, most ReID researches \cite{Shen2018PersonRW,Xu_2018_CVPR,Tay_2019_CVPR,Luo_2019_CVPR_Workshops,Jin_2020_CVPR} dedicate to conventional RGB-based cross-view matching tasks that use images captured by visible cameras in the daytime \cite{Ye2020CrossModalityPR}. However, poor illumination (e.g., at night-time) could degrade the recognition performance of visible cameras \cite{Liu2016TransferringDR}. With the widespread use of dual-mode cameras in recent years, infrared (IR) pedestrian images captured by IR imaging sensors, robust enough to illumination variations \cite{Wu2018CoupledDL}, have provided an effective and inexpensive complement to RGB images taken from visible-light cameras. Therefore, it becomes necessary to match IR person images with their RGB counterparts from disjoint cameras, which we call it the RGB-Infrared cross-modality person ReID (RGB-IR ReID) problem.

%%%%%%%%%% figure 1 %%%%%%%%%
\begin{figure}
\begin{center}
\includegraphics[width = 1\linewidth]
  {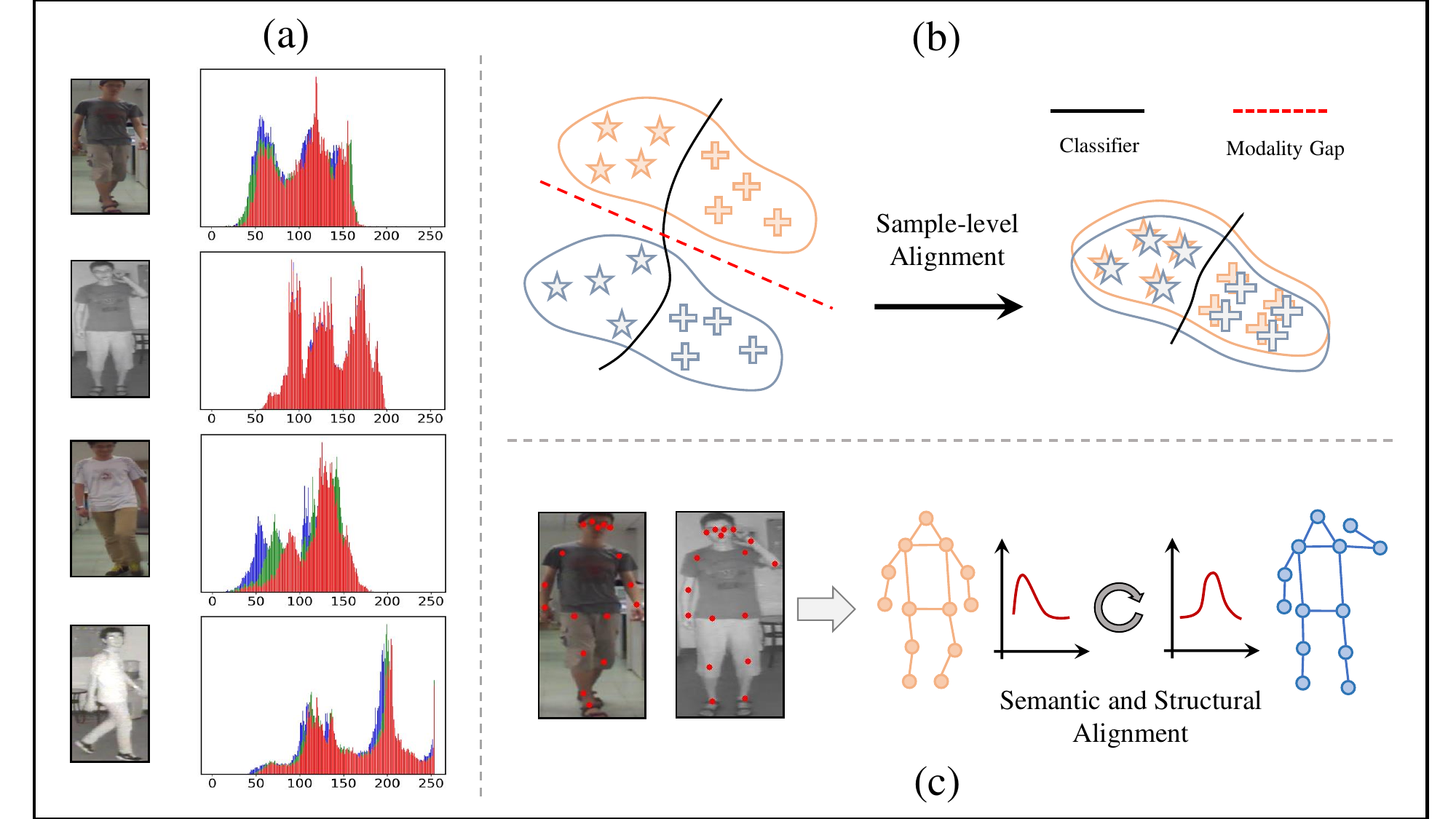}
\end{center}
\caption{(a) Histograms of Randomly selected RGB-IR images show that different persons present dissimilar distribution within each modality, while the same person shares similar distribution across modalities; (b) The proposed method operates on every cross-modality image pair and attempts to achieve precise sample-level alignment. (c) The semantic-aligned person-parts and body topology can be regarded as fine-grained cues to perform cross-modality alignment.}
\label{fig:1}
\end{figure}

Different from the visible spectrum-based re-identification problem, RGB-IR ReID encounters the extra modality discrepancy caused by different imaging mechanisms. RGB images store color information of three channels attained from visible light, whilst IR pictures contain only one channel information from infrared radiation \cite{Ma2019InfraredAV,Wang2020CrossModalityPG}. This intrinsic modality heterogeneity determines that existing single-modality ReID techniques cannot be directly used in cross-modality scenarios \cite{Huang2020ProbabilityWC}. For instance, although playing a critical role in RGB-based re-identification, color information becomes ineffective cues in RGB-IR ReID, since it is unable to find common visual areas with the same color in heterogeneous images \cite{Wu2020RGBIRPR}. Moreover, fundamental challenges in conventional ReID tasks, such as high inter-class similarity and dramatic intra-class variations caused by different illumination conditions, poses, viewpoints and backgrounds \cite{Ding2020MultitaskLW}, are further exacerbated by the modality gap, leading to more difficulties for matching tasks.

A crucial step in eliminating such modality gap is training deep networks to map heterogeneous pedestrian images to well-aligned representations that facilitate similarity-based matching tasks \cite{Ye2021DeepLF}. Currently, there mainly exists two lines of literature: (1) Pixel alignment-based methods \cite{Wang_2019_CVPR,Wang2019RGBInfraredCP,Wang2020CrossModalityPG,Choi_2020_CVPR}. They seek to alleviate the cross-modality variations from pixel level, often tending to adopt Generative Adversarial Network (GAN) to translate input images into their counterpart modality \cite{Wang_2019_CVPR,Wang2019RGBInfraredCP}. Nevertheless, adversarial training is unstable and easily suffers from the notorious mode collapse issue \cite{Kansal2020SDLSR,Ye2021VisibleInfraredPR}. (2) Feature alignment-based methods \cite{Ye2018VisibleTP,Ye2018HierarchicalDL,Dai2018CrossModalityPR,Hao2019HSMEHM,Zhang2019AttendTT,Lu_2020_CVPR,Ye2020BiDirectionalCT,Ye2020DynamicDA}. This line of researches aim to mitigate such heterogeneity with cross-modality feature alignment. In general, they first leverage modality-specific convolutional neural network (CNN) branches to map RGB and IR images into a common feature subspace, and then impose various discriminability constraints, such as triplet loss and contrastive loss, on the last layer of shared convolutional blocks, guiding the whole network to learn discriminative representations \cite{Ye2018HierarchicalDL,Ye2020BiDirectionalCT,Ye2021VisibleInfraredPR}. However, these point-/pair-wise similarity measures only consider the distance between RGB-IR samples rather than RGB-IR distributions, leaving the underlying distribution gap unresolved and thus impair modality-invariance of learned representations \cite{He2019WassersteinCL,Wang_2019_CVPR}. Although recent works employ probability measures such as Maximum Mean Discrepancy (MMD) \cite{Ye2020ImprovingNP} for modality alignment, they only enforce the alignment of global domain statistics, without considering statistical properties for each category \cite{Yang2020HeterogeneousGA}. Samples from different domains but the same class may still be aligned incorrectly \cite{Kang2019ContrastiveAN}.

To tackle above limitation, some current work focus their efforts on sample-level modality alignment approaches. They tend to employ global descriptors to minimize the distribution difference (Fig. \ref{fig:1} (a)) between cross-modality ReID sample pairs \cite{Wang2020CrossModalityPG}. This helps eliminate modality divergence within class, simultaneously improving the discriminability of learned features. However, some latent semantics and structural information (e.g., hair, face, and skeleton), which are significant clues for cross-modality alignment \cite{Wang2020DeepMM,Xu_2020_CVPR}, are ignored in these global-wise approaches. Although several recent studies leverage uniform partition strategy \cite{Wei2021FlexibleBP,Zhang2021MultiScaleCN} to decompose human body into parts for inter-modality discrepancy reduction, structural relations between body components are still not well exploited. Without the guidance of structural relations, the learned representations may only focus on local visual cues, failing to capture long-range dependencies among different body-parts.

In this paper, we propose a graph-enabled distribution matching solution, named Geometry-Guided Dual-Alignment (G$^2$DA), for RGB-IR ReID. Instead of solely attracting the set-level modality distributions, we present an adaptive skeleton graph to explicitly model high-order topology information of body parts. Thanks to the semantic-aligned nature of skeleton graph, we can jointly encourage semantic and structural modality consistency by matching the probabilistic distribution of global and part embeddings, forming a more robust feature subspace for cross-modal image retrieval. As depicted in Fig. \ref{fig:1} (c), we embed topological relations among body components into the skeleton graph, which contains not only person-dependent semantics but structural information, enabling dense correspondences establishment between RGB-IR sample pairs. Based on this formulation, we reframe modality-invariant feature learning as a graph matching problem, in which nodes and edges with the same semantic information are attracted to perform modality distribution matching. To solve this task, we introduce Optimal Transport (OT) \cite{Courty2017OptimalTF} as a matching metric to match semantically similar nodes. By minimizing the cost of a inter-modality transport plan, we could learn modality-invariant part representations while preserving their neighboring relations. Furthermore, we advance a Message Fusion Attention (MFA) mechanism to dynamically suppress noisy features caused by heavy occlusion, strengthening the discriminative power of learned representations. Extensive experiments on two public datasets validate the superiority of the proposed method, yielding competitive performance when compared with state-of-the-arts.

%%%%%%%%%%%%%%% contributions %%%%%%%%%%%%%%
Our major contributions are summarized as follows:
\begin{itemize}
\item We present a novel dual-alignment learning method for RGB-IR ReID, which jointly utilizes semantics and body topology to achieve precise sample-level modality alignment, effectively improving the performance of cross-modality retrieval.
\item We propose a G$^2$DA framework that embeds both semantic-aligned parts and structural relations of each sample into a skeleton graph, and formulates a OT-based graph matching task to jointly achieve semantic- and structural-level alignment, simultaneously enhancing feature invariance and discrimination.
\item A novel Message Fusion Attention (MFA) mechanism is advanced to adaptively refine the learned multi-scale features by semantic propagation, serving as an automated feature calibrator for discriminative feature learning. 
\item Extensive experimental results demonstrate the effectiveness of the proposed method over the state-of-the-arts on benchmark datasets.
\end{itemize}

%------------ 2. related work ---------------
%%%%%%%%%%%%%framework figure%%%%%%%%%%%%%%
\begin{figure*}
\begin{center}
\includegraphics[width = 1\linewidth]
{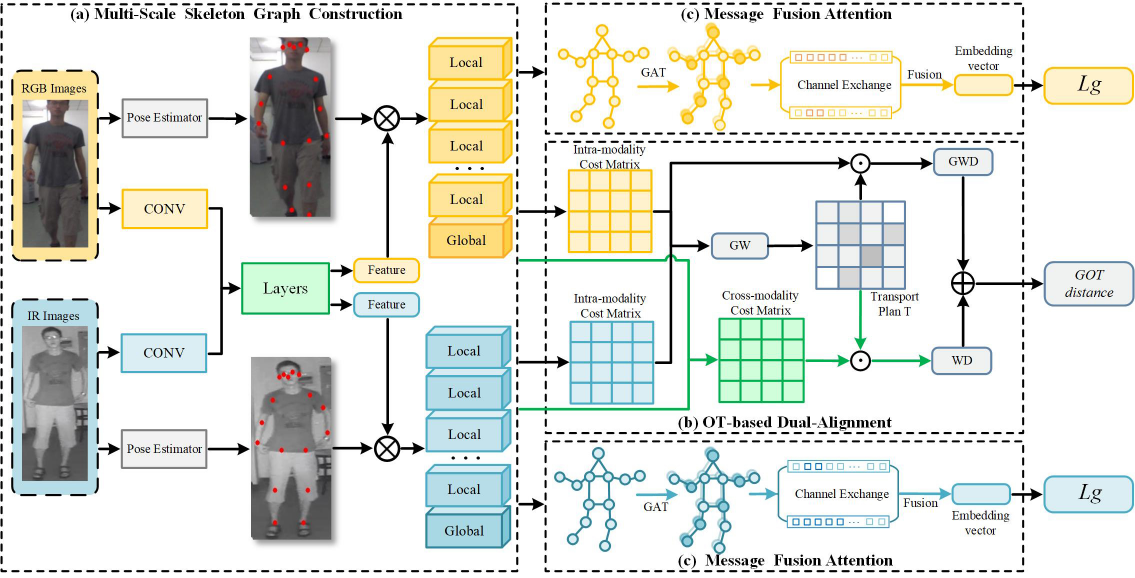}
\end{center}
\caption{Our G$^2$DA framework for RGB-IR person ReID. {\bf (a) Multi-Scale Skeleton Graph Construction} models the extracted whole-body contexts and local body-parts features into a skeleton graph; {\bf (b) OT-based Dual-Alignment} leverages both pair-wise semantics (\textit{Cross-modality Cost Matrix}) and group-wise structural similarities (\textit{Intra-modality Cost Matrix}) to perform sample-level modality alignment; {\bf (c) Message Fusion Attention (MFA)} enhances multi-head attention mechanism with a self-adaptive message fusion strategy to further boost feature discriminability.}
\label{fig:2}
\end{figure*}

\section{Related Work}
\subsection{RGB-IR Person ReID}
RGB-IR person ReID addresses the problem of matching a target person between heterogeneous modalities, which has witnessed remarkable progress \cite{Wu_2017_ICCV,Ye2018VisibleTP,Ye2018HierarchicalDL,Dai2018CrossModalityPR,Hao2019HSMEHM,Wang_2019_CVPR,Wang2019RGBInfraredCP,kang2019person,Zhang2019AttendTT,Tekeli_2019_ICCV,feng2019learning,Ye2019ModalityawareCL,Lu_2020_CVPR,Ye2020BiDirectionalCT,Ye2020DynamicDA,Zhu2020HeteroCenterLF,Kansal2020SDLSR,Wang2020CrossModalityPG,Choi_2020_CVPR,Ye2021VisibleInfraredPR} in recent years. Aside from intra-class variations, RGB-IR ReID has to deal with the additional modality discrepancy resulted from different wavelength ranges of RGB and IR imaging sensors. 
Wu \textit{et al.} \cite{Wu_2017_ICCV} contribute the first RGB-IR ReID dataset named SYSU-MM01, and present a zero-padding one-stream network for cross-modality matching. One commonly used deep architecture for cross-modality image matching is two-stream CNN network \cite{Ye2019ModalityawareCL}, where shallow layers are independent to learn a common feature space, and deep blocks are shared to mine discriminative cues for matching. In addition, a lot of metric learning techniques \cite{Ye2018HierarchicalDL,Ye2018VisibleTP,Ye2020BiDirectionalCT,Hao2019HSMEHM,Ye2020ImprovingNP,Zhang2019AttendTT} are proposed to yield more useful discriminative feature representations that's robust to heterogeneous modalities, achieving significantly better performance. Another avenues \cite{Wang_2019_CVPR,Wang2019RGBInfraredCP,Wang2020CrossModalityPG,Choi_2020_CVPR} adopt GAN-based image style transfer to remit pixel-level modality discrepancy. By virtue of this, Wang \textit{et al.} \cite{Wang2020CrossModalityPG} propose to generate cross-modality paired-images, and then utilize global descriptors to perform instance-level alignments. It greatly alleviates the issue of instance misalignment between modalities and produces better cross-modality matching results. Whereas the adversarial training process is unstable \cite{Liu2020ParametersSE}, and such fine-grained alignment seems to profit less from global features alone, leaving large room for further performance improvement.

\subsection{Local Feature Representation Learning}
To mine as much discriminative cues as possible, most ReID works focus their efforts on local feature learning. Horizontal division approaches \cite{Varior2016ASL,Sun2018BeyondPM} usually partition images into equal horizontal strips from top to bottom, while the pre-defined rigid grids are not well adapted to pose variations, imperfect pedestrian detectors and heavy occlusions \cite{Li_2017_CVPR,Ye2021DeepLF,BAI2020107036}. Human pose estimation techniques \cite{Insafutdinov2016DeeperCutAD,Sun_2019_CVPR} can alleviate the problem, but still suffer from noisy pose detection. More recently, there have emerged a lot of part-based methods in RGB-IR ReID \cite{Ye2020DynamicDA,Wang2020DeepMM,Wei2021FlexibleBP,Zhang2021MultiScaleCN}, in which most of them adopt horizontal division strategy incorporating attention mechanism to learn fine-grained modality-invariant features. This enables different-modality samples belonging to the same class to be tightly clustered in the learned feature subspace. Nevertheless, such partition strategy cannot guarantee the learned part features are semantically aligned across modalities, which might hinder dense correspondences learning between RGB-IR sample pairs.
% the crucial structural information (i.e., human-topology relations) has not been fully exploited for discriminative and invariant feature learning, which might incur a bottleneck of intra-class feature compactness. 
% lowering the upper bound of feature diversity.
% For better part-level alignment, our study leverages pose landmarks to extract accurate part features over human body, achieving a considerable boost in performance.

\subsection{Optimal Transport}
Optimal Transport (OT) distance \cite{kantorovich2006problem} is a classic metric that provides an effective way to measure the distance between probability distributions. Compared to Kullback-Leibler (KL) divergence or Jensen-Shannon (JS) divergence, OT is capable of measuring non-overlapping distributions, can be better adapted to gradient decent methods \cite{Liu2021OptimalTD}.nNowadays, it has been widely applied to various cross-domain tasks, including cross-domain alignment \cite{Chen2020GraphOT,Yuan2020WeaklySC,Chen2019UNITERLU} and unsupervised domain adaptation \cite{Li_2020_CVPR,Xu_2020_CVPR}. Through learning a transport plan ($T$) \cite{seguy2017large}, the cost \cite{Xu_2020_CVPR} of transferring feature distribution from one domain to another is minimized, i.e., the distribution divergence between domains is reduced \cite{Chen2020GraphOT}. Currently, several work also demonstrate that OT is able to preserve the geometry of feature distribution \cite{Courty2017OptimalTF,Damodaran_2018_ECCV,Lee2019HierarchicalOT}, showing promising potential to address identity recognition problems. Our work is the first to introduce OT into cross-modality ReID, which simultaneously measures point-wise distance and group-wise structural relations to achieve precise sample-level dual-alignment.

\subsection{Graph Attention Network}
Graph Attention Network (GAT) \cite{Velickovic2018GraphAN} is an attention-based architecture applied to graph-structured data. Compared with convolutional-based graph learning network (GCN) \cite{kipf2016semi}, GAT can evaluate the influence of different nodes on the target node \cite{huang2019signed}, so as to guide the model focus on the most task-relevant information \cite{Lee2018AttentionMI,Zhang_2021_CVPR}. It has been successfully adopted in person/vehicle ReID tasks \cite{Zhu2020ASG,bao2019masked,zhang2021person}. Most of them inject task-beneficial relation (e.g., extrinsic structured relation among images and inherent structured relation within an image \cite{Zhu2020ASG}) into the graph, and then learn node-wise similarities to refine original features. For cross-modality ReID, Ye \textit{et al.} \cite{Ye2020DynamicDA} pioneer a GAT-based feature learning module to reinforce global representations. The introduction of attention mechanism enables network to fully exploit cross-modality nuances for precise matching. In this work, we extend GAT with a self-adaptive message fusion strategy that reweights the information flow of semantic propagation, strengthening intra-modality part features for better dual-alignment.

%---------- 3. Proposed Method -----------
\section{Proposed Method}
This section details the proposed Geometry-Guided Dual-Alignment (G$^2$DA) framework (Fig. \ref{fig:2}). \textit{Multi-Scale Skeleton Graph Construction} (Sec. \ref{sec:1}) embeds deep semantics and structural relations of each sample into a skeleton graph, forming person-dependent and modality-independent representations for each identity. In \textit{OT-based Dual-Alignment} (Sec. \ref{sec:2}), we cast cross-modality alignment as a dense graph matching problem, introducing OT as a distribution metric to measure pair-wise distance and group-wise structural similarities between modalities. By learning a transport plan, divergences between RGB-IR graph pairs are as reduced as possible. To further suppress meaningless features caused by heavy occlusion, a \textit{Message Fusion Attention} (Sec. \ref{sec:3}) mechanism is presented as a feature calibrator for discriminative feature learning.

\subsection{Multi-Scale Skeleton Graph Construction}
\label{sec:1}
Unlike previous global-descriptor based approaches \cite{Wang2020CrossModalityPG}, we proposes a dynamic skeleton graph construction strategy that embeds multi-scale visual cues (i.e., body shape and local body-parts) of each identity into a complete graph. This graph provides hierarchical identifiable information for sample-level modality alignment.

The graph construction is a two-step process consisting of node and edge construction. We first adopt a two-stream CNN network \cite{Ye2020DynamicDA} as our baseline model for node representation learning. As shown in Fig. \ref{fig:2} (a), parameters of the first convolutional block are independent to map RGB and IR images into a common subspace, while the deeper four blocks are shared to mine discriminative features from that subspace. The global descriptor $\boldsymbol{V^{m}_{g}}$ ($m$ is the modality indicator) is obtained via Global Average Pooling (GAP) and a BNNeck \cite{Luo_2019_CVPR_Workshops}, which is optimized by the baseline loss $\mathcal{L}_{b}$:
%% global loss
\begin{equation}
\label{eq:baseline_loss}
\mathcal{L}_{b} = \mathcal{L}_{id} + \mathcal{L}_{tri},
\end{equation}
where softmax cross-entropy loss $\mathcal{L}_{id}$ is used for identity classification, and online hard-mining triplet loss $\mathcal{L}_{tri}$ \cite{Ye2020AugmentationIA} encourages the distance between positives to get closer while negatives farther. Note that positives (negatives) are sample pairs from different modalities.

To extract body-part features, we introduce a pose estimator \cite{Sun_2019_CVPR} to guide the feature learning process. Given an image, we can get its feature map $F_{cnn}\in \mathbb{R}^{2048\times H\times W}$ and thirteen key-point heatmap $F_{kp}\in \mathbb{R}^{H\times W}$ through the baseline model and pose estimator, respectively. For each $F_{kp}$, we duplicate it 2048 times and then perform outer product ($\otimes$) with $F_{cnn}$, where GAP operation ($g_{avg}(\cdot)$) is then applied to attain a group of key-points features $\boldsymbol{V^{m}_{l}}$:
\begin{equation}
\boldsymbol{V^{m}_{l}} = \{v^{m}_{k}\}^{K}_{k=1} = g_{avg}(F_{cnn} \otimes F_{kp}),
\label{eq:get_feature}
\end{equation}
where $K$ is the number of key-point, $v^{m}_{k} \in \mathbb{R}^{d}$ and $d$ is the feature dimension.

After obtaining these multi-scale features, each image is represented as a set of feature vectors ($\boldsymbol{V^{m}_{g}}$ and $\boldsymbol{V^{m}_{l}}$). Considering the high-order topology information of body-parts, we then connect any two possible vectors for edge representation learning. In this way, two types of relationships are explicitly modeled by edges. First, each edge between adjacent semantic nodes reflects the local-local relation, i.e., the body topology. Second, every local part is connected with a whole-body embedding. The edges between them reflect the global-local relation where global features offer informative contextual guidance for local cues. Thus, we generate a complete graph, in which nodes refer to extracted whole-part features and edges indicate structural similarities among them.

%%%%%%%%%%%%% keypoint figure %%%%%%%%%%%%%%
\begin{figure}[t]
\begin{center}
\includegraphics[width=1\linewidth]
{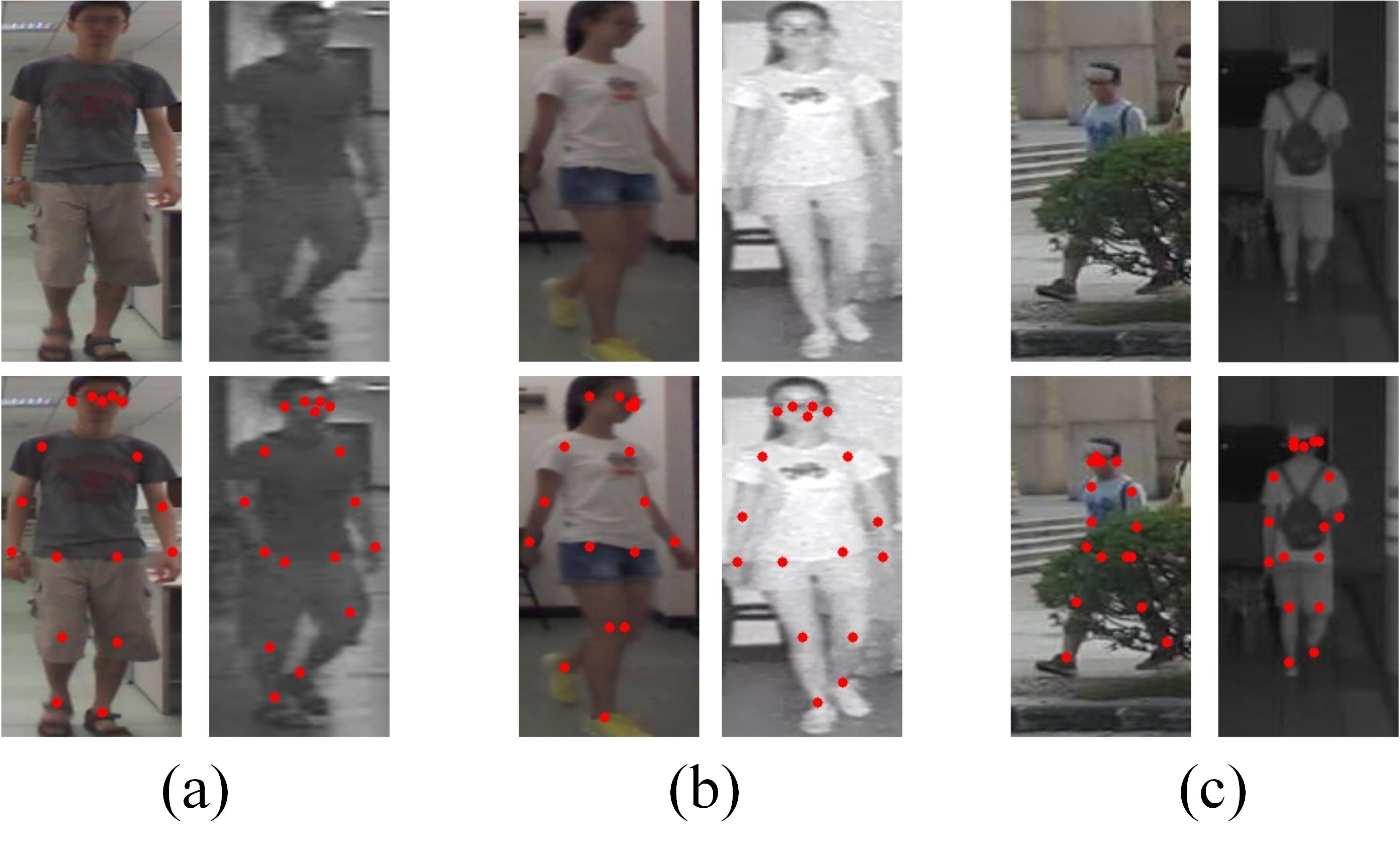}
\end{center}
\caption{Visualization results of pose estimation in our cross-modality settting. (a) and (b) verify the pose estimator can work well on both RGB and IR images, while (c) indicates that occlusions can contaminate the learned part features.}
\label{fig:keypoint}
\end{figure}

{\bf Discussion.} Regarding the extracted pose-guided part features, one question is likely to come up: the pose estimator we present is typically trained with RGB samples, whether it works on IR images? Or could our pose estimator correctly fight against the modality gap? To answer this question, we examine the visualization results of estimated key-points in both RGB and IR images (Fig. \ref{fig:keypoint}) and observe that our pose estimator can work effectively in the cross-modal settings. One possible reason is that the prediction process of human landmarks relies more on modality-shared information such as body structure and shape, rather than modality-specific cues like color \cite{abdellali2019robust}. But in some cases, the learned features are contaminated by occlusions (Fig. \ref{fig:keypoint} (c)). In Sec. \ref{sec:3}, we present a strategy that helps alleviate this limitation via emphasizing identity-related parts while depressing noisy ones.

%%%%% OT-based Modality Alignment %%%%%%
\subsection{OT-based Dual-Alignment}
\label{sec:2}
As discussed in Introduction section, either contrastive loss \cite{Ye2018HierarchicalDL} or triplet loss \cite{Li2020XModality,Lu_2020_CVPR,Ye2021VisibleInfraredPR,Ye2020DynamicDA} cannot well eliminate the distribution divergence between RGB-IR modalities \cite{Shen2018WassersteinDG}. To bridge such gap, Ye \textit{et al.} \cite{Ye2020ImprovingNP} propose to employ a distribution metric to enforce alignment between the entire RGB and IR sets. Yet the distribution variances within each class are ignored, possibly leading to misalignment within the same class. Here, we suggest a sample-level dual-alignment method for simultaneously easing set- and sample-level misalignment. Our dual-alignment approach comprehensively considers semantic and structural information: on the one side, we cluster skeleton graph nodes with the same semantics across modalities to attain modality-agnostic information. On the other side, we measure node-wise similarities (i.e., edges of the graph) and encourage these structural relations to be consistent between modalities, further enhancing feature invariance in a relational manner.

Given a RGB-IR skeleton graph pair, $G_1 = (V_1,E_1)$ and $G_2 = (V_2,E_2)$ with $|V_1| = |V_2| = K+1$ ($V$ and $E$ denotes the set of nodes and edges respectively), dual-alignment aims to establish an assignment between $G_1$ and $G_2$ by maximizing a graph matching score over corresponding nodes and edges \cite{Zanfir2018DeepLO}. To this end, the optimal assignment $\mathbf{v}^{*}$ can be formulated as: 
\begin{equation}
\mathbf{v}^{*}=\underset{\mathbf{v}}{\operatorname{argmax}} \mathbf{v}^{\top} \mathbf{M} \mathbf{v}, \text { s.t. }\|\mathbf{v}\|_{2}=1
\label{eq:3}
\end{equation}
where $\mathbf{v} \in \mathbb{R}^{(K+1)(K+1)\times 1}$ measures how well every point $i \in V_1$ matches with $a \in V_2$, and positive matrix $\mathbf{M} \in \mathbb{R}^{(K+1)(K+1)\times (K+1)(K+1)}$ measures those of every pair $(i,j) \in E_1$ and $(a,b) \in E_2$.

To solve Eq. (\ref{eq:3}), we need to simultaneously preserve the distribution consistency of semantic-aligned nodes as well as their structural relationships (i.e., edges) between modalities. However, commonly used distribution measures, such as Kullback-Leibler or Jensen-Shannon divergence, are unable to evaluate the underlying geometric properties of the feature subspace, thus easily falling into local optimum \cite{Arjovsky2017WassersteinGA, Liu2021OptimalTD,Xu_2020_CVPR}. Alternatively, as a structured distribution measure, Optimal Transport (OT) is capable of modeling geometric displacements across multi-modal images, exhibiting a great advantage in cross-domain alignment \cite{Chen2019UNITERLU,Lee2019HierarchicalOT}. In our work, we use OT to learn a transport plan $T$ \cite{seguy2017large} that transfers the unit quality from RGB (IR) to IR (RGB) modality with minimal cost. Specifically, a Graph Optimal Transport (GOT) distance \cite{Chen2020GraphOT} is employed as the matching metric for the transport plan learning. Benefiting from the combination of Wasserstein distance (WD) \cite{Peyr2019ComputationalOT} and Gromov-Wasserstein distance (GWD) \cite{Peyr2016GromovWassersteinAO} in GOT, we can jointly measure pair-wise relations and group-wise structural similarities to facilitate the dual-alignment in shared feature subspace.

{\bf Semantic-level alignment.}
We leverage the ability of measuring point-wise distance in WD to match two groups of semantic features cross heterogeneous modalities. Let $\mathbb{S}^{R}$ and $\mathbb{S}^{I}$ be complete metric spaces (e.g., Euclidean space), $\mu\in P(\mathbb{S}^{R})$ and $\nu\in P(\mathbb{S}^{I})$ denote two discrete probability distributions, $\prod(\mu,\nu)$ denote the space of joint probability distributions with marginals $\mu$ and $\nu$. Given two sets of feature vectors $V^{R}=V^{R}_{l} \cup V^{R}_{g}$ and $V^{I}=V^{I}_{l} \cup V^{I}_{g}$ associated with a same identity $y_{i}$ from $\mathbb{S}^{R}$ and $\mathbb{S}^{I}$ respectively, we use WD to promote alignment between two sets of embeddings. The transport plan $T$ seeks a joint probability distributions $\gamma\in\prod(\mu,\nu)$ that yields a minimal displacement cost:
\begin{equation}
\begin{aligned}
D_{w}(\mu,\nu) 
& = \inf_{\gamma \in \prod(\mu,\nu)} \mathbb{E}_{(v^{R},v^{I})\sim \gamma}[c(v^{R},v^{I})] \\ 
& = \min_{T \in \prod(u,v)} \sum_{i=1}^{K+1} \sum_{j=1}^{K+1} T_{ij} \cdot c(v^{R}_{i},v^{I}_{j}),
\label{eq:WD}
\end{aligned}
\end{equation}
here $\!\prod(u,v)\!\!=\!\!\{T\!\!\in\!\!\mathbb{R}^{(K\!+\!1)\times(K\!+\!1)}\!\!\mid\!\! T1_{K+1}\!\!=\!\!u,T^\top1_{K+1}\!\!=\!\!v\}$ (vector $1_{K+1}=[1,...,1]^\top \in \mathbb{R}^{K+1}$), $v^{R}_{i}\in V^{R}$ and $v^{I}_{j}\in V^{I}$. The weight vectors $u=\{u_{i}\}^{K+1}_{i} \in \Delta _{K+1}$ and $v=\{v_{i}\}^{K+1}_{i} \in \Delta _{K+1}$ are ($K$+$1$)-dimensional simplex. $c(\cdot,\cdot)$ is the cost function (e.g., cosine similarity) that measures the distance between two groups of feature $V^{R}$ and $V^{I}$, forming a \textit{Cross-Modality Cost Matrix} as seen in Fig. \ref{fig:2}(b).

We further impose contrastive loss \cite{Hadsell2006DimensionalityRB} over every pair of aligned features, which enforces intra-class feature compactness whereas inter-class separability. Different from previous methods, the contrastive loss is applied to both local body-part features and global features, where every pair-wise features is of different modalities.
\begin{equation}
L_{c}\hspace{-0.1cm}=\hspace{-0.1cm}\frac{1}{2N_{t}(K\hspace{-0.1cm}+\hspace{-0.1cm}1)} \sum_{k=1}^{K+1} \sum_{n=1}^{N_{t}} (1-y_{n}) d^{2}_{n,k}+ y_{n} \{\max (\varepsilon-d_{n,k}, 0)\}^{2},
\label{eq:CL}
\end{equation}
here $N_{t}$ is the training batch size, and $d_{n,k}=\|v^{R}_{n,k}-v^{I}_{n,k}\|_{2}$. $y_{n}$ is a binary label indicating whether two images belong to the same identity, $y_{n}=0$ means they are paired images and $y_{n}=1$ otherwise. $\varepsilon$ is a pre-defined margin, which is set to 2.0 in our experiments. Only if $y_{n}=1$ and their distance is smaller than $\varepsilon$, the latter term will be incurred.

{\bf Structure-level alignment.}
To preserve the structural relationships of semantic nodes across heterogeneous modalities, we calculate distances between key-points within each modality and compare these distances in RGB(IR) graph with their counterparts by GWD \cite{Xu2019ScalableGL}.

More concretely, in each modality, we calculate the similarity scores between node embeddings (the \textit{Intra-Modality Cost Matrix} shown in the Fig. \ref{fig:2}). The scores can be viewed as the edge between two nodes, and edge alignment encourages both local and global relations across two modalities to be consistent in the common subspace. Following the definition in Eq. (\ref{eq:WD}), we calculate the Gromov-Wasserstein distance as follows:
\begin{equation}
    \begin{split}
        D_{gw}(\mu\!,\!\nu)\!&=\!\!\inf_{\gamma \in\!\prod(\mu,\nu)}\! \mathbb{E}_{(V^{R}\!,\!V^{I})\sim \gamma, (V^{R^\prime}\!,\!V^{I^\prime})\sim \gamma}
        [\mathcal{L}\!(\!V^{R}\!,\!V^{I}\!,\!V^{R^\prime}\!,\!V^{I^\prime}\!)] \\ 
        &= \min_{T \in \prod(u,v)} \sum_{i,i^\prime,j,j^\prime} T_{ij} T_{i^\prime j^\prime} \mathcal{L}(v^{R}_{i},v^{I}_{j},v^{R^\prime}_{i},v^{I^\prime}_{j}),
    \label{eq:GWD}
    \end{split}
\end{equation}
where $\mathcal{L}(\cdot)$ is the cost function used to calculate the structural similarity between two pairs of features $(v^{R}_{j},v^{R^\prime}_{j})$ and $(v^{I}_{j},v^{I^\prime}_{j})$, formulated as: $\mathcal{L}(v^{R}_{i},v^{I}_{j},v^{R^\prime}_{i},v^{I^\prime}_{j})=\| c(v^{R}_{i},v^{R^\prime}_{i})-c(v^{I}_{j},v^{I^\prime}_{j}) \|$. Notably, the learned transport plan $T$ for GWD is shared with WD, so that both semantic and structural information can be exploited to achieve more precise cross-modality alignment. The unified learning object is formulated as:
\begin{equation}
     D_{ot}(\mu,\nu) = \min_{T \in \prod(u,v)} [\phi D_{w}(\mu,\nu) + (1-\phi) D_{gw}(\mu,\nu)],
\label{eq:GOT}
\end{equation}
where $\phi$ is a hyper-parameter to control the importance of two distances. In this way, the distribution discrepancy between cross-modality sample pairs can be further alleviated by enforcing such dual-alignment.

%% Message Fusion Attention %%
\subsection{Message Fusion Attention}
\label{sec:3}
Although Fig. \ref{fig:keypoint} (a) (b) demonstrates the pose estimator functions well under both RGB and IR modalities, noisy pose detections caused by heavy occlusion still exist (Fig. \ref{fig:keypoint} (c)). To address this issue, we advance a novel Message Fusion Attention (MFA) module, which, in an adaptive manner, emphasizes identity-related part features while suppressing the meaningless ones (i.e., part features of occluded regions), effectively enhancing feature expressiveness. Specifically, for each image, MFA refines the extracted skeleton graph with natural adjacency of different body parts and performs multi-head self-attention among all parts. To fuse knowledge learned from different heads, MFA introduces a parameter-free message fusion strategy, acting as a soft gate to adaptively control what information should be fused.

{\bf Graph Refinement.} 
We refine the extracted skeleton graph into an undirected version $G = (V,E)$ based on natural body-part adjacency, in which $V$ represents node set $ V^{m} = V^{m}_{l} \cup V^{m}_{g}$ ($m \in \{R,I\}$) and $E$ is a set of edges that formulates an adjacency matrix $A \in \{0,1\}^{(K+1)\times(K+1)}$. $A_{i,j} = 1$ means that the $i$-th and the $j$-th parts are connected and 0 otherwise. Notably, aside from local-local relations, we also establish global-local relations by connecting global descriptor with all of local components. In this way, global features provide context guidance (e.g., body shape, gender and clothes) to improve local features \cite{Wang2020DeepMM}, and local details, in turn, complement the global descriptor with dispersive fine-grained identity information (e.g., hair, face and feet) \cite{Yang_2019_CVPR}.

{\bf Intra-Graph Attention.} 
In $G = (V,E)$, each node is allowed to attend over its connected nodes, thus learning a set of parameters (i.e., the attention scores) to aggregate its neighbors for node updating. Rather than a single attention function, we perform multi-head attention on each node \cite{Vaswani2017AttentionIA} to capture richer relevant information within each graph. It allows different body regions (e.g., upper body and lower body) to be parallelly attended by distinct heads, thereby raising the upper bound of feature distinctiveness for better retrieval results.

Given the input node embeddings $V_{k}=\{v_{k}\in\mathbb{R}^{d}\} ^{K+1}_{k=1}$ ($d$ is the feature dimension), we define a learnable attention coefficient between every pair of nodes to evaluate their similarities. Assume that there are $L$ attention heads, for the $l$-th attention head, the attention coefficients between the $i$-th and the $j$-th node $\alpha^{l}_{i,j}$ can be formulated as:
\begin{equation}
 \alpha^{l}_{i,j} = \frac{exp(\Gamma(\vec{a}^{l}[W \cdot v_{i} \parallel W \cdot v_{j}]))}{\sum\limits_{\forall A_{i,k}=1} exp(\Gamma(\vec{a}^{l}[W \cdot v_{i} \parallel W \cdot v_{k}]))},
\label{eq:ac}
\end{equation}
where $\Gamma(\cdot)$ denotes the LeakyReLU activation function, $\vec{a}^{l} \in \mathbb{R}^{2d}$ is a learnable weight vector of attention mechanism $a^{l}$, $W$ is a weight matrix used for feature dimension transformation, and $\parallel$ indicates the concatenation operation. For the $i$-th node, we calculate the attention coefficients with all of its neighbors for aggregation, resulting in a partial-representation of the $l$-th head:
\begin{equation}
    \displaystyle v_{i,l} = \sigma (\sum_{\forall A_{i,j}=1} \alpha^{l}_{i,j} \cdot W^{l} \cdot v_{j}),
\end{equation}
where $\sigma(\cdot)$ is a nonlinear activation function, and $W^{l}$ is the corresponding input linear transformation’s weight matrix. During the training process, the message passing of semantic nodes is promoted whilst that of meaningless ones is suppressed adaptively.

{\bf Self-Adaptive Message Fusion.} 
Even with rich knowledge learned from multiple distinct subspaces, there is no guarantee that all attained information is useful. Here, we introduce a parameter-free message fusion strategy, which enables redundant channels to be adaptively replaced by a meaningful one to augment the expressive ability of each partial-representation. Specifically, this process is self-guided by a channel-wise scaling factor, which is calculated by Batch Normalization (BN) operation applied to each head:
\begin{equation}
    \displaystyle v_{i,l,c}^{\prime}=\gamma_{i,l,c} \frac{v_{i,l,c}-\mu_{i,l,c}}{\sqrt{\sigma_{i,l,c}^{2}+\epsilon}}+\beta_{i,l,c},
\label{eq:BN}
\end{equation}
where the subscript $c$ denotes the $c$-th channel; $\mu_{i,l,c}$ and $\sigma_{i,l,c}$ are the mean and standard deviation of the $c$-th channel for all the $L$ attention heads; $\gamma_{i,l,c}$ and $\beta_{i,l,c}$ are the learnable scaling factor and offset, respectively; $\epsilon$ is a small constant that used to avoid zero denominator.

Note that, the scaling factor $\gamma_{i,l,c}$, which evaluates the importance of the $c$-th channel, determines whether $v_{i,l,c}^{\prime}$ is potentially redundant or not \cite{Wang2020DeepMF}. Guided by this factor, the process of self-adaptive message fusion can be formulated as:
\begin{equation}
    v_{i,l,c}^{\prime}=\left\{\begin{array}{ll}
    \displaystyle\gamma_{i,l,c} \frac{v_{i,l,c}-\mu_{i,l,c}}{\sqrt{\sigma_{i,l,c}^{2}+\epsilon}}+\beta_{i,l,c}, \qquad \qquad \text{if } \gamma_{i, l, c}>\theta 
    \\
    \\
    \displaystyle\frac{1}{L-1} \sum_{l^{\prime} \neq l}^{L} \gamma_{i,l^{\prime},c} \frac{v_{i,l^{\prime},c}-\mu_{i,l^{\prime},c}}{\sqrt{\sigma_{i,l^{\prime},c}^{2}+\epsilon}}+\beta_{i,l^{\prime},c}, \quad \text{otherwise}
\end{array}\right.
\label{eq:MF}
\end{equation}
where $\theta$ is a pre-defined threshold. A channel should be replaced by the mean of other channels if its scaling factor $\gamma_{i,l,c}$ is smaller than $\theta$. Once all the enhanced partial-representations $v^{\prime}_{i,l}$ are obtained, a better output node embedding $\boldsymbol{\overline{v}_{i}}$ can be returned via a simple concatenation operation.

{\bf Aggregation Layer.} 
To obtain a unified embedding, we aggregate the enhanced partial-representations in the following way:
\begin{equation}
V_{G} = BN(g_{avg}(F_{cnn})) + \sum_{k=1}^{K+1} \omega_{k} \boldsymbol{\overline{v}_{k}},
\label{eq:graph}
\end{equation}
here, $V_{G}$ is the output graph representation, which is optimized by cross-entropy $\mathcal{L}_{g}$ to ensure intra-modality discrimination. $BN(\cdot)$ is the Batch Normalization operation. The first term is utilized to stabilize the training process. And in the second term, we employ a learnable weight vector $\omega_{k}$ to aggregate node embeddings, where meaningless ones are expected to get small weight.

%%%%%%%%% Train and Test %%%%%%%%
\subsection{Network Training and Inference}
During training, we aim to minimize the following objective functions:
\begin{equation}
    \mathcal{L} = \lambda_{b}\mathcal{L}_{b} + \lambda_{o}D_{ot}(\mu,\nu) + \lambda_{c} L_{c} + \lambda_{g}\mathcal{L}_{g},
    \label{eq:all}
\end{equation}
where $\lambda_{b}$, $\lambda_{o}$, $\lambda_{c}$ and $\lambda_{g}$ are trade-off parameters that indicate the weight of each loss function.

In the testing phase, only \textit{Multi-Scale Skeleton Graph Construction} and \textit{Message Fusion Attention} are adopted, making the inference pipeline computationally efficient and robust. The final graph embedding $V_{G}$ is utilized to perform re-identification. 
                  
%%%%%%%%%%%%% SYSU-MM01 dataset %%%%%%%%%%%%%
\linespread{1.1}
\begin{center}
\begin{table*}[!htbp]
\Huge
\caption{Comparison with the state-of-the-arts on SYSU-MM01 dataset under two search modes. Accuracy (\%) at Rank r and mAP are reported.}
\label{table1}
\resizebox{\textwidth}{!}{
\begin{tabular}{ccccccccccccccccc}
\Xhline{1.9pt}
\multicolumn{1}{c}{\multirow{3}{*}{Method}}    & \multicolumn{8}{!{\vrule width1.9pt}c!{\vrule width1.9pt}}{All-search}                 & \multicolumn{8}{c}{Indoor-search}    \\ \Xcline{2-17}{1.9pt} 
 & \multicolumn{4}{!{\vrule width1.9pt}c}{Single-shot} & \multicolumn{4}{c!{\vrule width1.9pt}}{Multi-shot} & \multicolumn{4}{c}{Single-shot} & \multicolumn{4}{c}{Multi-shot} \\ \Xcline{2-17}{1.9pt} 
\multicolumn{1}{c}{}                        &\multicolumn{1}{!{\vrule width1.9pt}c}{r1}     & \multicolumn{1}{c}{r10}    & \multicolumn{1}{c}{r20}   & \multicolumn{1}{c!{\vrule width1.9pt}}{mAP}   & \multicolumn{1}{c}{r1}     & \multicolumn{1}{c}{r10}   & \multicolumn{1}{c}{r20}   & \multicolumn{1}{c!{\vrule width1.9pt}}{mAP}   & \multicolumn{1}{c}{r1}     & \multicolumn{1}{c}{r10}    & \multicolumn{1}{c}{r20}   & \multicolumn{1}{c!{\vrule width1.9pt}}{mAP}   & \multicolumn{1}{c}{r1}     & \multicolumn{1}{c}{r10}   & \multicolumn{1}{c}{r20}   & mAP   \\ \Xhline{1.9pt}
\multicolumn{1}{c}{HOG \cite{dalal2005histograms}}                     & \multicolumn{1}{!{\vrule width1.9pt}c}{2.76}   & 18.3   & 31.9  & \multicolumn{1}{c!{\vrule width1.9pt}}{4.24}  & 3.82   & 22.8  & 37.6  & \multicolumn{1}{c!{\vrule width1.9pt}}{2.16}  & 3.22   & 24.7   & 44.5  & \multicolumn{1}{c!{\vrule width1.9pt}}{7.25}  & 4.75   & 29.2  & 49.4  & 3.51  \\
\multicolumn{1}{c}{LOMO \cite{liao2015person}}                    & \multicolumn{1}{!{\vrule width1.9pt}c}{3.64}   & 23.2   & 37.3  & \multicolumn{1}{c!{\vrule width1.9pt}}{4.53}  & 4.70   & 28.2  & 43.1  & \multicolumn{1}{c!{\vrule width1.9pt}}{2.28}  & 5.75   & 34.4   & 54.9  & \multicolumn{1}{c!{\vrule width1.9pt}}{10.2}  & 7.36   & 40.4  & 60.3  & 5.64  \\
\multicolumn{1}{c}{Zero-Padding \cite{Wu_2017_ICCV}}            & \multicolumn{1}{!{\vrule width1.9pt}c}{14.8}   & 54.1   & 71.3  & \multicolumn{1}{c!{\vrule width1.9pt}}{15.9}  & 19.1   & 61.4  & 78.4  & \multicolumn{1}{c!{\vrule width1.9pt}}{10.9}  & 20.6   & 68.4   & 85.8  & \multicolumn{1}{c!{\vrule width1.9pt}}{26.9}  & 24.4   & 75.9  & 91.3  & 18.6  \\
\multicolumn{1}{c}{TONE+HCML \cite{Ye2018HierarchicalDL}}               & \multicolumn{1}{!{\vrule width1.9pt}c}{14.3}   & 53.2   & 69.2  & \multicolumn{1}{c!{\vrule width1.9pt}}{16.2}  & -      & -     & -     & \multicolumn{1}{c!{\vrule width1.9pt}}{-}     & -      & -      & -     & \multicolumn{1}{c!{\vrule width1.9pt}}{-}     & -      & -     & -     & -     \\
\multicolumn{1}{c}{BDTR \cite{Ye2018VisibleTP}}                    & \multicolumn{1}{!{\vrule width1.9pt}c}{17.0}   & 55.4   & 72.0  & \multicolumn{1}{c!{\vrule width1.9pt}}{19.7}  & -      & -     & -     & \multicolumn{1}{c!{\vrule width1.9pt}}{-}     & -      & -      & -     & \multicolumn{1}{c!{\vrule width1.9pt}}{-}     & -      & -     & -     & -     \\
\multicolumn{1}{c}{D-HSME \cite{Hao2019HSMEHM}}                  & \multicolumn{1}{!{\vrule width1.9pt}c}{20.7}   & 62.8   & 78.0  & \multicolumn{1}{c!{\vrule width1.9pt}}{23.2}  & -      & -     & -     & \multicolumn{1}{c!{\vrule width1.9pt}}{-}     & -      & -      & -     & \multicolumn{1}{c!{\vrule width1.9pt}}{-}     & -      & -     & -     & -     \\
\multicolumn{1}{c}{IPVT+MSR \cite{kang2019person}}                & \multicolumn{1}{!{\vrule width1.9pt}c}{23.2}   & 51.2   & 61.7  & \multicolumn{1}{c!{\vrule width1.9pt}}{22.5}  & -      & -     & -     & \multicolumn{1}{c!{\vrule width1.9pt}}{-}     & -      & -      & -     & \multicolumn{1}{c!{\vrule width1.9pt}}{-}     & -      & -     & -     & -     \\
\multicolumn{1}{c}{cmGAN \cite{Dai2018CrossModalityPR}}                   & \multicolumn{1}{!{\vrule width1.9pt}c}{27.0}   & 67.5   & 80.6  & \multicolumn{1}{c!{\vrule width1.9pt}}{27.8}  & 31.5   & 72.7  & 85.0  & \multicolumn{1}{c!{\vrule width1.9pt}}{22.3}  & 31.6   & 77.2   & 89.2  & \multicolumn{1}{c!{\vrule width1.9pt}}{42.2}  & 37.0   & 80.9  & 92.1  & 32.8  \\
\multicolumn{1}{c}{D$^{2}$RL \cite{Wang_2019_CVPR}}                     & \multicolumn{1}{!{\vrule width1.9pt}c}{28.9}   & 70.6   & 82.4  & \multicolumn{1}{c!{\vrule width1.9pt}}{29.2}  & -      & -     & -     & \multicolumn{1}{c!{\vrule width1.9pt}}{-}     & -      & -      & -     & \multicolumn{1}{c!{\vrule width1.9pt}}{-}     & -      & -     & -     & -     \\
\multicolumn{1}{c}{DGD+MSR \cite{feng2019learning}}                 & \multicolumn{1}{!{\vrule width1.9pt}c}{37.4}   & 83.4   & 93.3  & \multicolumn{1}{c!{\vrule width1.9pt}}{38.1}  & 43.9   & 86.9  & 95.7  & \multicolumn{1}{c!{\vrule width1.9pt}}{30.5}  & 39.6   & 89.3   & 97.7  & \multicolumn{1}{c!{\vrule width1.9pt}}{50.9}  & 46.6   & 93.6  & 98.8  & 40.1  \\
\multicolumn{1}{c}{JSIA-ReID \cite{Wang2020CrossModalityPG}}               & \multicolumn{1}{!{\vrule width1.9pt}c}{38.1}   & 80.7   & 89.9  & \multicolumn{1}{c!{\vrule width1.9pt}}{36.9}  & 45.1   & 85.7  & 93.8  & \multicolumn{1}{c!{\vrule width1.9pt}}{29.5}  & 43.8   & 86.2   & 94.2  & \multicolumn{1}{c!{\vrule width1.9pt}}{52.9}  & 52.7   & 91.1  & 96.4  & 42.7  \\
\multicolumn{1}{c}{AlignGAN \cite{Wang2019RGBInfraredCP}}                & \multicolumn{1}{!{\vrule width1.9pt}c}{42.4}   & 85.0   & 93.7  & \multicolumn{1}{c!{\vrule width1.9pt}}{40.7}  & 51.5   & 89.4  & 95.7  & \multicolumn{1}{c!{\vrule width1.9pt}}{33.9}  & 45.9   & 87.6   & 94.4  & \multicolumn{1}{c!{\vrule width1.9pt}}{54.3}  & 57.1   & 92.7  & 97.4  & 45.3  \\
\multicolumn{1}{c}{AGW \cite{Ye2021DeepLF}}                & \multicolumn{1}{!{\vrule width1.9pt}c}{47.5}   & 84.39   & 92.14  & \multicolumn{1}{c!{\vrule width1.9pt}}{47.65}  & -   & -  & -  & \multicolumn{1}{c!{\vrule width1.9pt}}{-}  & 54.17   & 91.14   & 95.98  & \multicolumn{1}{c!{\vrule width1.9pt}}{62.97}  & -   & -  & -  & -  \\
\multicolumn{1}{c}{Xmodal \cite{Li2020XModality}}                & \multicolumn{1}{!{\vrule width1.9pt}c}{49.92}   & 89.79   & 95.96  & \multicolumn{1}{c!{\vrule width1.9pt}}{50.73}  & -   & -  & -  & \multicolumn{1}{c!{\vrule width1.9pt}}{-}  & -   & -   & -  & \multicolumn{1}{c!{\vrule width1.9pt}}{-}  & -   & -  & -  & -  \\
\multicolumn{1}{c}{DDAG \cite{Ye2020DynamicDA}}                & \multicolumn{1}{!{\vrule width1.9pt}c}{54.75}   & 90.39   & 95.81  & \multicolumn{1}{c!{\vrule width1.9pt}}{53.02}  & 61.83   & 92.68  & 97.49  & \multicolumn{1}{c!{\vrule width1.9pt}}{47.06}  & 61.02   & 94.06  & \textbf{98.41}  & \multicolumn{1}{c!{\vrule width1.9pt}}{67.98}  & 69.23   & 95.13   & 98.31 & 59.42  \\
\Xhline{1.9pt}
\multicolumn{1}{c}{Ours}                & \multicolumn{1}{!{\vrule width1.9pt}c}{\textbf{57.07}}   & \textbf{90.99}   & \textbf{96.28}  & \multicolumn{1}{c!{\vrule width1.9pt}}{\textbf{55.05}}  & \textbf{64.47}   & \textbf{94.22}  & \textbf{98.18}  & \multicolumn{1}{c!{\vrule width1.9pt}}{\textbf{48.91}}  & \textbf{63.70}   & \textbf{94.06}   & 98.35  & \multicolumn{1}{c!{\vrule width1.9pt}}{\textbf{69.83}}  & \textbf{72.62}   & \textbf{95.99}  & \textbf{98.87}  & \textbf{61.89}  \\
\Xhline{1.9pt} 
\end{tabular}}
\end{table*}
\end{center}
%%%%%%%%%%%%%%%%%%%%%%%%%%%%%%%%%%%%%%%%%%%%

\begin{figure}[t]
    \centering
    \includegraphics[width=0.9\linewidth]{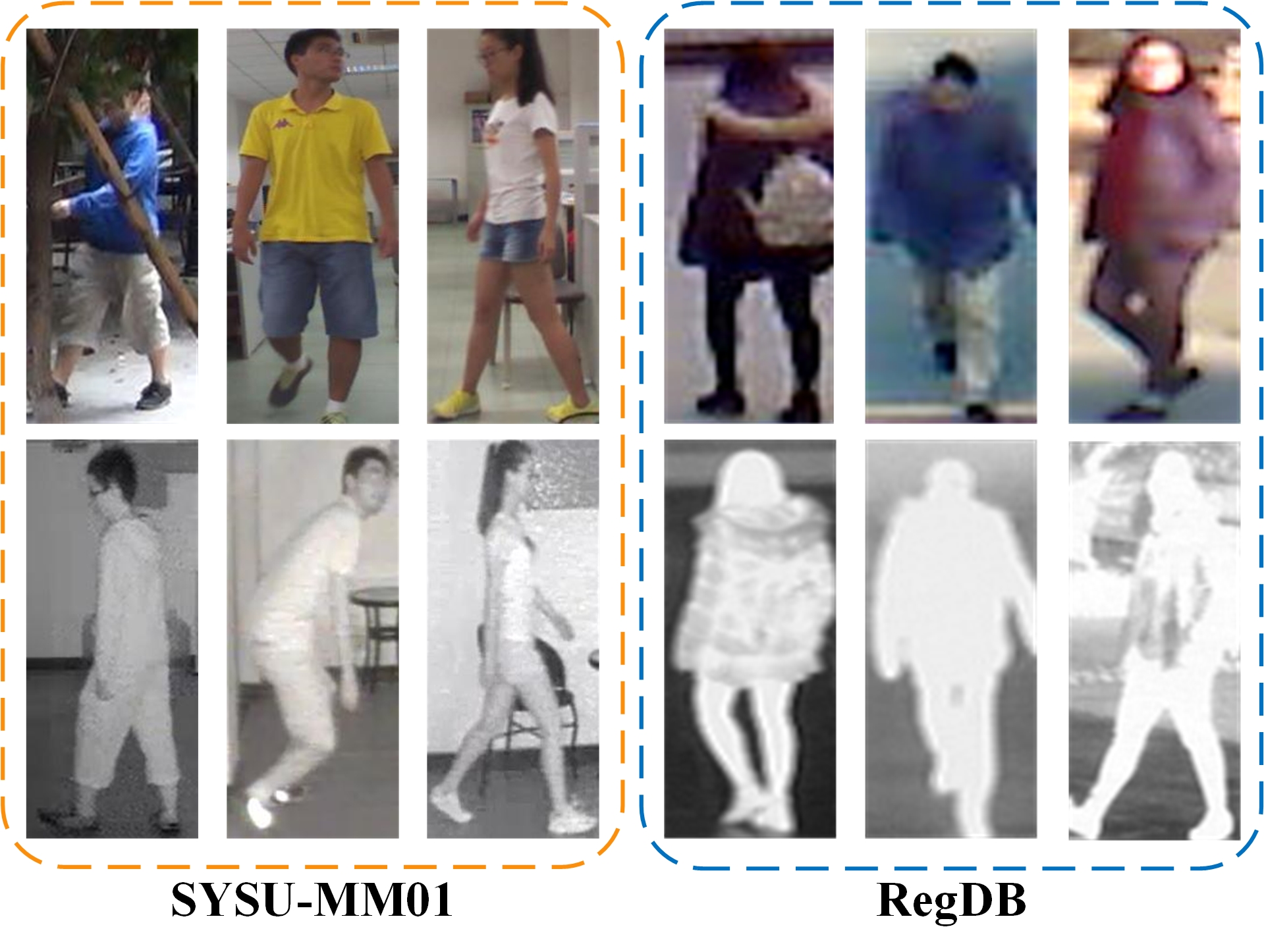}
    \caption{Example images randomly sampled from SYSU-MM01 \cite{Wu2020RGBIRPR} and RegDB \cite{Nguyen2017PersonRS} datasets. The top row shows RGB images all captured by visible cameras, for IR images in the bottom row, the front half is taken by near-infrared cameras while the rest is by far-infrared cameras.}
\label{fig:dataset}
\end{figure}

%------------ 4. EXPERIMENT ---------------
\section{Experiment}
\label{sec:experiment}
We evaluate the proposed method on two cross-modality ReID datasets (SYSU-MM01 and RegDB), performing fair comparison with state-of-the-art methods. The ablation study also carried out to validate the effects of each component.

\subsection{Experimental Settings}
{\bf Datasets.}
SYSU-MM01 \cite{Wu2020RGBIRPR} is a large-scale RGB-IR dataset containing RGB and IR images of 491 identities captured by four visible and two near-infrared cameras under both indoor and outdoor scenes. We adopt the fixed split that using 395 identities with 22,258 RGB images and 11,909 IR images for training and 96 identities for testing. Following the evaluation protocol \cite{Wu2020RGBIRPR}, there are two search modes, i.e., \textit{all-search} and \textit{indoor-search}. The former contains images captured by all four visible cameras and the latter only comprises images captured by two indoor visible cameras. For each mode, we adopt both \textit{single-shot} and \textit{multi-shot} settings, which means that the gallery set contains one/ten randomly selected RGB image(s) for each IR query image. It is worth noting that \textit {single-shot \& all-search} is the most challenging test mode.

RegDB \cite{Nguyen2017PersonRS} is a relatively small-scale dataset collected by a dual-camera system (one visible and one far-infrared). We present some samples from two datasets to illustrate the difference of IR images captured by near-infrared and far-infrared cameras (Fig. \ref{fig:dataset}). RegDB accommodates 8,240 images of 412 identities, where each identity has 10 RGB and 10 IR images, respectively. We randomly divide the dataset into two halves, one for training and the other for testing, and repeated a total of 10 times random splits of the gallery and probe sets \cite{Ye2018HierarchicalDL}. During the test stage, both \textit{Visible-to-Thermal} and \textit{Thermal-to-Visible} evaluation modes are adopted.

{\bf Evaluation Protocols.}
To evaluate the performance of our model, Cumulative Matching Characteristic (CMC) curve and mean Average Precision (mAP) are adopted as the evaluation criteria. The CMC at rank-k represents the probability of the correct match occurs in top-k retrieval results, and mAP is a comprehensive evaluation measuring the retrieval performance of multi-matching. Different from the conventional ReID task, during the testing stage, the query set and gallery set are of different modalities in cross-modality tasks.

{\bf Implementation details.}
The proposed method is developed on Pytorch framework with NVIDIA 2080Ti GPU. Following existing RGB-IR ReID works \cite{Ye2020DynamicDA,Ye2021DeepLF}, we adopt a two-stream CNN network with ResNet50 backbone for fair comparison \cite{He2016DeepRL}, and apply the parameters pre-trained on ImageNet \cite{Deng2009ImageNetAL} for network initialization. Specifically, the parameters are different in the first block, while deep convolutional blocks are shared for each modality. As for the pose estimation model, we adopt the state-of-the-art HR-Net \cite{Sun_2019_CVPR} pre-trained on the COCO dataset \cite{Lin2014MicrosoftCC} and get $K=13$ key-points including head, shoulders, elbows, wrists, hips, knees, and ankles, following \cite{Wang2020HighOrderIM}. Multi-head attention mechanism is adopted to optimize the similarity relationships of part features, where the number of attention heads $L$ is set to 4.

Both in training and testing stage, the input images are firstly resized to $288 \times 144$. Random cropping with zero-padding and horizontal flipping are adopted for data augmentation \cite{Ye2020DynamicDA}. The batch size $N_t$ is set to 64, and each identity comprises 4 RGB and 4 IR images, for a total of 8 identities. We set the learning rate as 0.1 for both datasets, which is decayed by 0.1 at 30-th epoch and 0.01 at 50-th epoch. Our model is trained with total 80 epochs and is optimized with stochastic gradient descent (SGD), where momentum parameter is set to 0.9. By default, we set the margin parameter $\rho$ in triplet loss to 0.3, $\varepsilon$ in contrastive loss to 2.0 and $\theta$ in channel exchange to 0.02, the hyper-parameter $\phi$ in Eq. (\ref{eq:GOT}) to 0.5. The trade-off parameter $\lambda_{b}$ = $\lambda_{o}$ = $\lambda_{g}$ = 1, and $\lambda_{c}$ = 0.1.

\subsection{Comparison With the State-of-the-Arts}
In this subsection, we conduct a thorough comparison of the proposed method with both conventional and state-of-the-art methods, including HOG \cite{dalal2005histograms}, LOMO \cite{liao2015person}, Zero-Padding \cite{Wu_2017_ICCV}, TONE+HCML \cite{Ye2018HierarchicalDL}, BDTR \cite{Ye2018VisibleTP}, D-HSME \cite{Hao2019HSMEHM}, IPVT+MSR \cite{kang2019person}, cmGAN \cite{Dai2018CrossModalityPR}, D$^2$RL \cite{Wang_2019_CVPR}, DGD+MSR \cite{feng2019learning}, JSIA-ReID \cite{Wang2020CrossModalityPG}, AlignGAN \cite{Wang2019RGBInfraredCP}, AGW \cite{Ye2021DeepLF}, Xmodal \cite{Li2020XModality} and DDAG \cite{Ye2020DynamicDA}.

{\bf Results on SYSU-MM01 dataset.}
We adopt both \textit {all-search} and \textit {indoor-search} test modes, and each mode is performed in both \textit {single-shot} and \textit {multi-shot} ways. As shown in Table \ref{table1}, our method outperforms state-of-the-art approaches on the most challenging \textit {single-shot \& all-search} test mode, obtaining 57.50\% Rank-1 accuracy and 55.91\% mAP scores. Through in-depth analysis, we make the observations as follows: (1) Hand-craft features (HOG, LOMO) cannot well be generalized to the cross-modality matching task; (2) Feature learning with instance-level constraints, such as classification loss, triplet loss and ranking loss, shows significant improvement in performance. Whereas, these learning objectives mainly concentrate on optimizing the relationship between RGB-IR samples, the distribution divergence between two modalities has not been properly solved, which can degenerate the re-identification performance; (3) Adversarial learning-based methods (cmGAN, D$^2$RL, JSIA-ReID, AlignGAN) also achieve high re-identification accuracy. Most of them leverage GAN to generate cross-modality images for the reduction of pixel-level modality discrepancy, but they complicate the inference pipeline by appending a pre-processing module, and the adversarial training is unstable \cite{Ye2020DynamicDA}. (4) Recent shared feature learning methods (AGW, DDAG) significantly outperform the state-of-the-art GAN-based method (AlignGAN), which also show considerable efficiency as they are free of adversarial training. Compared to these methods, our work explicitly utilizes the crucial visual properties for each identity, and simultaneously performs semantic- and structural-level matching to achieve precise sample-level modality alignment. The superiority of such fine-grained modality alignment has been validated by the reported experimental results.

%%%%%%%%% retrieval results %%%%%%%%%%
\begin{figure*}[!htbp]
    \centering
    \includegraphics[width=1\linewidth]
    {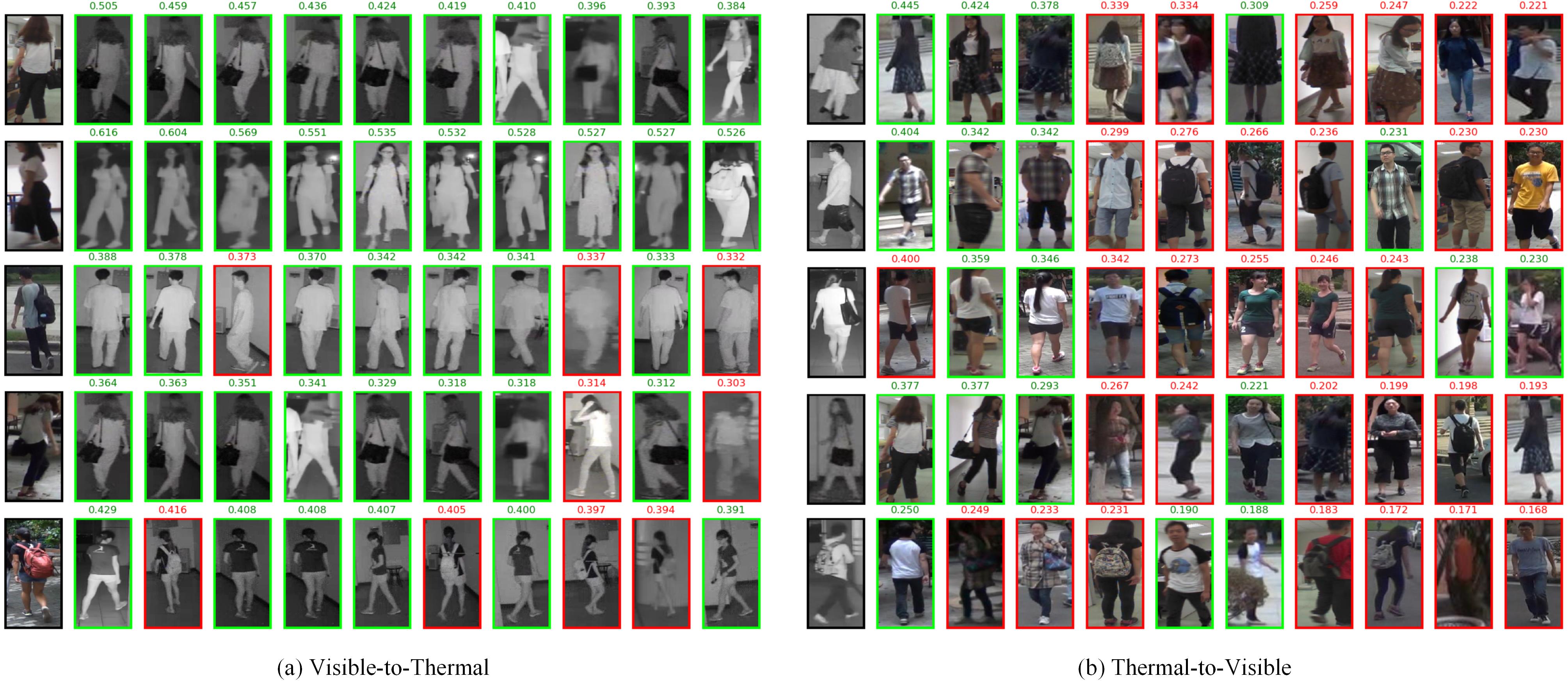}
    \caption{Top-10 retrieval results of 10 randomly selected identities on SYSU-MM01 in \textit{Visible-to-Thermal} and \textit{Thermal-to-Visible} query modes. Green and Red boxes denote correct and wrong matching results, respectively. The value on top of each image represents its cosine similarity to the query image.}
\label{fig:5}
\end{figure*} 

%%%%%%%%%%%%%%%%% RegDB %%%%%%%%%%%%%%%%%%%%%
\renewcommand\arraystretch{0.9}
\begin{table*}[h]
\begin{center}
\caption{Comparison with the state-of-the-arts on RegDB dataset with two test modes. Accuracy (\%) at Rank r and mAP are reported.}
\label{table:regdb}
\linespread{2}
\resizebox{0.85\textwidth}{!}{
\begin{tabular}{c|cccc|cccc}
\hline
\multirow{2}{*}{Method} & \multicolumn{4}{c|}{Visible to Thermal}  & \multicolumn{4}{c}{Thermal to Visible} \\ \cline{2-9} 
                        &\multicolumn{1}{c}{r1}        & \multicolumn{1}{c}{r10}        & \multicolumn{1}{c}{r20}        & \multicolumn{1}{c|}{mAP}        & \multicolumn{1}{c}{r1}         & \multicolumn{1}{c}{r10}        & \multicolumn{1}{c}{r20}        & \multicolumn{1}{c}{mAP}       \\ \hline
Zero-Padding \cite{Wu_2017_ICCV}          & 17.75      & 34.21      & 44.35      & 18.90      & 16.63      & 34.68      & 44.25      & 17.82     \\
TONE + HCML \cite{Ye2018HierarchicalDL}            & 24.44      & 47.53      & 56.78      & 20.88      & 21.70      & 45.02      & 55.58      & 22.24     \\
BDTR \cite{Ye2018VisibleTP}                    & 33.56      & 58.61      & 67.43      & 32.76      & 32.92      & 58.46      & 68.43      & 31.96     \\
D$^{2}$RL \cite{Wang_2019_CVPR}                  & 43.40       & 66.10       & 76.30       & 44.10       & -          & -          & -          & -         \\
DGD+MSR \cite{feng2019learning}               & 48.43      & 70.32      & 79.95      & 48.67      & -          & -          & -          & -         \\
JSIA-ReID \cite{Wang2020CrossModalityPG}               & 48.10       & -          & -          & 48.90       & 48.50       & -          & -          & 49.30      \\
D-HSME \cite{Hao2019HSMEHM}                  & 50.85      & 73.36      & 81.66      & 47.00      & 50.15      & 72.40      & 81.07      & 46.16     \\
IPVT+MSR \cite{kang2019person}                & 58.76      & 85.75      & 90.27      & 47.85      & -          & -          & -          & -         \\
AlignGAN \cite{Wang2019RGBInfraredCP}                & 57.90       & -          & -          & 53.60       & 56.30       & -          & -          & 53.40      \\
Xmodal \cite{Li2020XModality}               & 62.21      & 83.13      & 91.72      & 60.18      & -          & -          & -          & -         \\
AGW \cite{Ye2021DeepLF}               & 70.05      & -     & -      & \textbf{66.37}     & -          & -          & -          & -         \\
DDAG \cite{Ye2020DynamicDA}                    & 69.34      & 86.19      & 91.49      & 63.46      & 68.06      & \textbf{85.15}    &\textbf{90.31}     & 61.80     \\ \hline
\textbf{Ours}                    &\textbf{71.72}            &\textbf{87.13} &\textbf{91.92}           & 65.90           &\textbf{69.50}            &84.87   &   89.85         &\textbf{63.88}  \\ \hline
\end{tabular}}
\end{center}
\end{table*}
%%%%%%%%%%%%%%%%%%%%%%%%%%%%%

{\bf Retrieval Results Visualization.}
To evaluate the cross-modality retrieval capability of the proposed method, we visualize the top-10 image retrieval results for five randomly selected samples in \textit{Thermal-to-Visible} and \textit{Visible-to-Thermal} query settings (Fig. \ref{fig:5}). We observe that our ReID model performs well even with the `back-view of person' query conditions (e.g., row \textbf{a}, \textbf{c}, \textbf{d} and \textbf{f} in Fig. \ref{fig:5}) in cross-modality settings. This might benefit from the dual-alignment operations performed on each class, where the combination of human body parts and topology ensure these heterogeneous samples of different class can be better recognized. As we also can see from Fig. \ref{fig:5}), our model achieves better performance in the \textit{Visible-to-Thermal} query mode than in the  \textit{Thermal-to-Visible} mode, because visible images contain more visual cues that are conducive to matching than IR images. From row \textbf{c}, \textbf{e} and \textbf{j} in Fig. \ref{fig:5}, we surprisingly see that our method is still able to output correct Rank-1 results when coping with the situation that personal belongings (e.g., backpacks) changes. This consequence probably profits from the hierarchical skeleton graph mechanism, which enables our model be more focused on human body rather than identity-unrelated cues. On the whole, the Visualized retrieval results in both query modes further prove that our proposed method effectively augments the modality-invariance and discriminability in learned representations, thereby yielding impressive retrieval accuracy.

{\bf Results on RegDB dataset.}
Table \ref{table:regdb} presents the comparison results on RegDB dataset, where both \textit{Visible-to-Thermal} and \textit{Thermal-to-Visible} test modes are adopted. It is clear that our method achieves comparable performance to existing approaches, attaining 71.57\% and 69.11\% in terms of Rank-1 identification rate under two test modes, respectively, which further verifies that our model is robust to heterogeneous query settings. From the perspective of test mode, the performance of \textit{Visible-to-Thermal} is better than that of \textit{Thermal-to-Visible}, possibly benefiting from rich visual cues containing in visible images.

%%%%%%%%% Model Analysis %%%%%%%%%%%%%%
\subsection{In-Depth Model Analysis}
{\bf Ablation Study.}
To verify the contribution of each component to the model, we conduct ablation experiments on standard benchmark SYSU-MM01 dataset with both \textit{all-search} and \textit{indoor-search} modes, and both modes are performed in \textit{single-shot} manner. To be specific, ‘‘$B$’’ denotes the baseline model, a partially shared two-stream CNN network trained by $L_{b}$. ‘‘$O$’’ indicates OT-based dual-alignment. ‘‘$CL$’’ is the contrastive loss. ‘‘$G$’’ indicates intra-graph attention and ‘‘$MF$’’ represents self-adaptive message fusion. The results are reported in Table \ref{tab:3}.

%%%%%%%%%%%%% ablation %%%%%%%%%%%
\linespread{1.2}
\begin{table*}[!t]
\caption{Ablation study on SYSU-MM01 dataset under two search modes. Accuracy (\%) at Rank r and mAP are reported.}
\begin{center}
\resizebox{0.8\textwidth}{!}{
\begin{tabular}{l!{\vrule width0.6pt}cccc!{\vrule width0.6pt}cccc}
\Xhline{0.6pt}
Search modes       & \multicolumn{4}{c!{\vrule width0.6pt}}{All-search} & \multicolumn{4}{c}{Indoor-search} \\ \Xhline{0.6pt}
Methods       & r1      & r10      & r20      & mAP     & r1       & r10       & r20      & mAP      \\ \Xhline{0.6pt}
$B$            &  49.17  &   84.38 &     92.16&  47.27     & 53.60      &    89.67    &   95.63  &   61.41  \\ \Xcline{1-1}{0.6pt}
$B$+$O$         &     53.58          &   88.26    &      94.27    &    52.00     &     57.13         &     92.23     &       97.55   &   64.86   \\ \Xcline{1-1}{0.6pt}
$B$+$O$+$CL$   &      54.56       &     88.89     &     95.05    &     53.52     &   57.74             &      93.47    &  98.09   &  65.69      \\ \Xcline{1-1}{0.6pt}
$B$+$O$+$CL$+$G$    &     56.65       &     90.84     &    96.18      &    54.83     &    61.80           &    93.85      &    97.93 & 68.77      \\ \Xcline{1-1}{0.6pt}
$B$+$O$+$CL$+$G$+$MF$ &      57.07       &      90.99    &      96.28    &     55.05    &      63.70           &    94.06      &        98.35 &   69.83  \\ \Xhline{0.6pt} 
\end{tabular}
}
\end{center}
\label{tab:3}
\end{table*}
%%%%%%%%%%%%%%%%%%%%%%%%%%%%%%%%%%%%%%%%

We adopt a two-stream CNN network following \cite{Ye2020DynamicDA} as our baseline model, where the first residual block is independent for each modality to capture modality-specific feature patterns and the other four blocks are shared to learn modality-shared feature embeddings. The combination of identity loss and triplet loss guides the network to learn cross-modality discriminative feature representations. It achieves 49.17\% and 47.27\% in terms of Rank-1 accuracy and mAP score, respectively, outperforming a lot of previous methods and proving to be a strong baseline model.

\textit{Effectiveness of O.}
Based on $B$, we further employ OT distance to explicitly mitigate the distribution difference between every cross-modality sample pair, which exhibits a clear improvement over the baseline model by 4.41\% on Rank-1 accuracy and 4.73\% on mAP score under the challenging \textit {all-search} mode (a larger improvement can be seen in \textit {indoor-search} mode). It is obvious that the bridge of sample-level distribution gap greatly benefits modality-invariant feature learning, effectively facilitating cross-modality matching.

\textit{Effectiveness of CL.}
Based on $B$+$O$, we study the effectiveness of contrastive loss and observe that the performance get a further improved by 0.98\% on Rank-1 identification rate and 1.52\% on mAP score. The above phenomenon shows the importance of part-level feature alignment, and simultaneously verifies that contrastive loss is conducive to  enhance modality-invariance of learned representations.

\textit{Effectiveness of G.}
When we further perform multi-head attention within the graph, the matching performance is advanced with gains of 2.09\% and 4.06\% in Rank-1 accuracy under \textit{all-search} and \textit{indoor-search} modes, respectively. It is evident that richer knowledge captured from distinct subspaces greatly raises the upper bound of feature distinctiveness, effectively facilitating discriminative feature representation learning.

\textit{Effectiveness of MF.}
We further introduce a self-adaptive message fusion strategy to refine partial-representations learned from multi-head attention mechanism. As we can see, it acquires 1.9\% performance gain of on Rank-1 identification rate and 1.06\% improvement on mAP score under \textit{indoor-search} mode, demonstrating that replacing the redundant channels with a meaningful one is beneficial to learn high-quality representations.

It is obvious that the proposed model reaches the optimal performance when integrating with all components, obtaining 57.07\% on Rank-1 accuracy and 55.05\% on mAP score, which suggests that all these modules work well together.

%%%%%%%%%%%%%%% t-SNE %%%%%%%%%%%%%%%%%
\begin{figure}[t]
    \centering
    \includegraphics[width=1\linewidth]{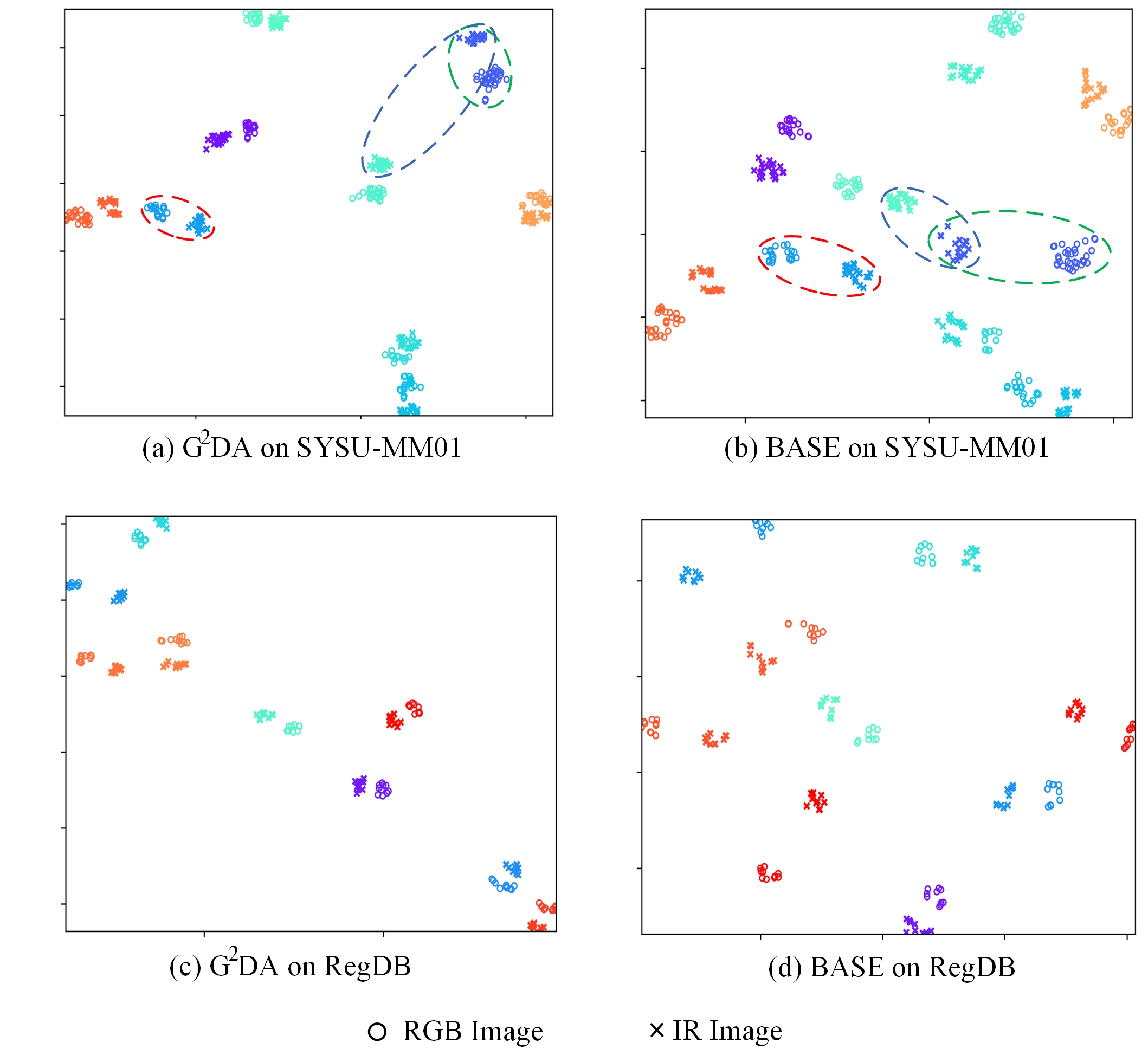}
    \caption{Feature distribution comparisons of \textbf{dual-alignment method \textit{vs}. the baseline model} on SYSU-MM01 (a)(b) and RegDB (c)(d) using \textit{t}-SNE visualization.
    Circle and cross shapes represent sample's modality, and each color stands for an identity.}
    \label{fig:6}
\end{figure}
%%%%%%%%%%%%%%%%%%%%%%%%

%%%%%%%%%%%%%%% parameter %%%%%%%%%%%%%%%%%
\begin{figure*}[t]
    \centering
    \includegraphics[width=1\linewidth]
    {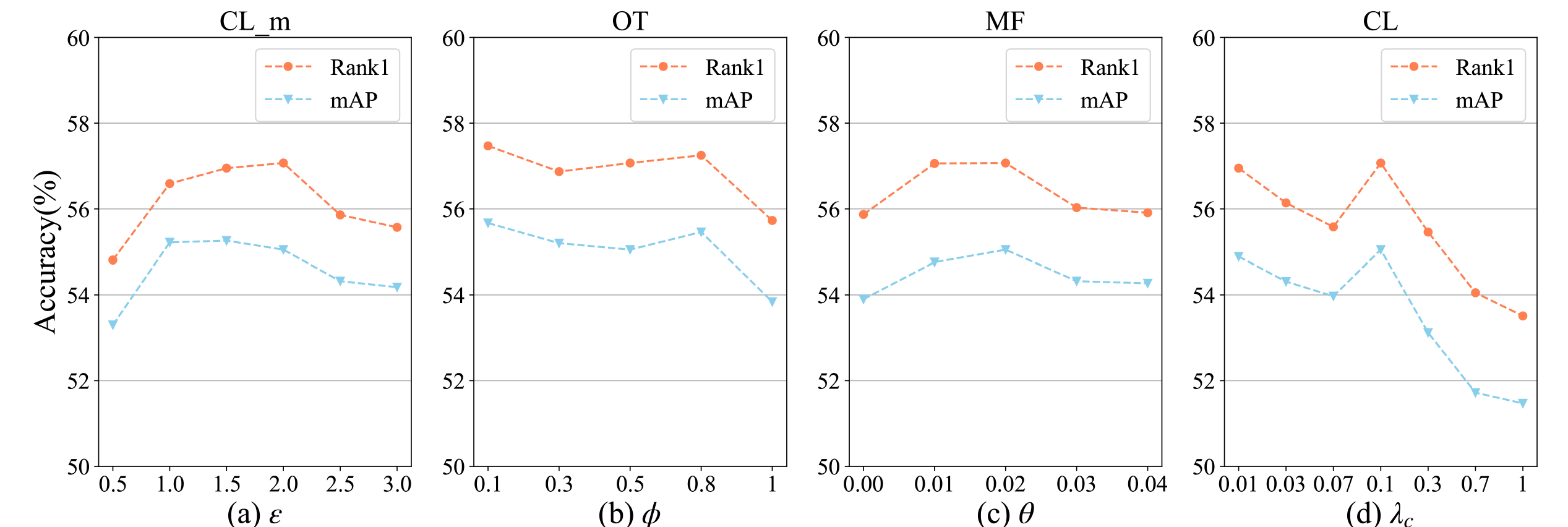}
    \caption{Evaluation of the margin $\varepsilon$ of contrastive loss in Eq. (\ref{eq:CL}), the hyper-parameter $\phi$ in Eq. (\ref{eq:GOT}), the threshold $\theta$ of channel exchange in Eq. (\ref{eq:MF}) and the trade-off parameter $\lambda_{c}$ in Eq. (\ref{eq:all}) on SYSU-MM01 dataset with \textit {single-shot} $\&$ \textit{all-search} test mode. Rank-1 accuracy (\%) and mAP score(\%) are reported. Best viewed in color.}
\label{fig:7}
\end{figure*}
%%%%%%%%%%%%%%%%%%%%%%%%%%

{\bf Feature Distribution Comparisons with Baseline.}
To further demonstrate the feature learning effectiveness of our sample-level dual-alignment mechanism, we illustrate the feature distributions learned by G$^2$DA and the baseline in Fig. \ref{fig:6} using \textit{t}-distributed Stochastic Neighbor Embedding (\textit{t}-SNE) \cite{Maaten2008VisualizingDU}. Here we show the visualization results of feature embeddings of different-modality images that belong to nine different identities (randomly sampled on SYSU-MM01 and RegDB, respectively), which are extracted by our dual-alignment model and the baseline network.

Comparing the region enclosed by the red dashed ellipse in Fig. \ref{fig:6} (a) with its counterparts in Fig. \ref{fig:6} (b), we see that the dual-alignment approach considerably reduces intra-class variations in cross-modality settings. The phenomenon that inter-class class distance is less than intra-class distance (e.g., comparing the regions enclosed by the blue and green dashed ellipses in Fig. \ref{fig:6} (b)) has also been alleviated by G$^2$DA (see their counterparts in Fig. \ref{fig:6} (a)). Overall, The dual-alignment method outperforms the baseline in improving intra-class compactness and inter-class separation, either on SYSU-MM01 (Fig. \ref{fig:6} (a)(c)) or RegDB (Fig. \ref{fig:6} (b)(d)).

{\bf Parameters Analysis.}
This subsection evaluates the effect of four key parameters in the proposed method, i.e., the margin $\varepsilon$ of contrastive loss in Eq. (\ref{eq:CL}), the hyper-parameter $\phi$ in Eq. (\ref{eq:GOT}), the threshold $\theta$ of channel exchange in Eq. (\ref{eq:MF}) and the trade-off parameter $\lambda_{o}$ in Eq. (\ref{eq:all}). We conduct all the experiments on SYSU-MM01 dataset with the most challenging \textit {single-shot \& all-search} test mode. The results are presented in Fig. \ref{fig:7}.

(1) The margin $\varepsilon$ is utilized to define the distance difference between negative sample pairs. We vary $\varepsilon$ from 0.5 to 3 and evaluate the performance under these settings. It can be observed that either a smaller or a larger $\varepsilon$ would impair the performance, and $\varepsilon \in [1.5,2]$ can achieve better performance.

(2) The hyper-parameter $\phi$ is used to control the trade-off between WD and GWD. We observe that the re-identification performance could be consistently improved when $\phi \in [0.1,0.8]$, but drops dramatically when only WD works (i.e., $\phi = 1$). Results in Fig. \ref{fig:7} (b) demonstrate that both WD and GWD are crucial for modality alignment.

(3) The threshold $\theta$ is designed to determine whether a channel should be exchanged. We investigate the trend of performance when $\theta$ ranges from 0 to 0.04 ($\theta=0$ means that the channel exchange module is removed). It can be seen that both Rank-1 accuracy and mAP score are consistently improved with increasing $\theta$, and then decrease after $\theta = 0.02$.

(4) The trade-off parameter $\lambda_{c}$ indicates the influence of contrastive loss to the retrieval performance, and a proper selection of $\lambda_{c}$ is of critical importance. As depicted in Fig. \ref{fig:7} (d), we vary $\lambda_{c}$ from 0.01 to 1, and observe that this term could improve the cross-modality ReID accuracy when $\lambda_{c} = 0.1$. This benefit is brought by the further enhancement of intra-class compactness and inter-class separability.

Empirically, we set $\varepsilon = 2$, $\phi = 0.5$, $\theta = 0.02$ and $\lambda_{c} = 0.1$ in all our experiments. Notably, better performance might be achieved by different parameter selection.

\section{Conclusion}
In this paper, we present a graph-enabled distribution matching solution, named Geometry-Guided Dual-Alignment (G$^2$DA), for RGB-IR ReID. Based on a pose-adaptive skeleton graph generator, G$^2$DA involves a dual-alignment mechanism, i.e., semantic and structure alignments, to jointly strengthen the invariance and discriminability of learned representations. Specifically, we equip the deep network with semantic-aligned prior of body part locations to map  heterogeneous images into complete skeleton graphs, in which both high-level semantics and structural information are embedded for representation learning. This novel formulation translates modality alignment into a sample-level graph matching problem, aiming to discover modality correspondence by considering both node features and edge attributes. To this end, we introduce OT distance as a matching metric and learn a modality-invariant feature subspace by minimizing the cost of a inter-modality transport plan. To enhance model robustness against noisy features, we further propose a MFA module to adaptively refine the learned features by promoting semantic message fusion, effectively augmenting the expressiveness of semantic-aligned node embeddings. Extensive experimental results on two standard benchmark datasets validate the effectiveness of G$^2$DA, with consistent improvement over previous state-of-the-arts.

\bibliographystyle{IEEEtran}
\bibliography{IEEEabrv,reference}

\begin{thebibliography}{10}
\providecommand{\url}[1]{#1}
\csname url@samestyle\endcsname
\providecommand{\newblock}{\relax}
\providecommand{\bibinfo}[2]{#2}
\providecommand{\BIBentrySTDinterwordspacing}{\spaceskip=0pt\relax}
\providecommand{\BIBentryALTinterwordstretchfactor}{4}
\providecommand{\BIBentryALTinterwordspacing}{\spaceskip=\fontdimen2\font plus
\BIBentryALTinterwordstretchfactor\fontdimen3\font minus
  \fontdimen4\font\relax}
\providecommand{\BIBforeignlanguage}[2]{{%
\expandafter\ifx\csname l@#1\endcsname\relax
\typeout{** WARNING: IEEEtran.bst: No hyphenation pattern has been}%
\typeout{** loaded for the language `#1'. Using the pattern for}%
\typeout{** the default language instead.}%
\else
\language=\csname l@#1\endcsname
\fi
#2}}
\providecommand{\BIBdecl}{\relax}
\BIBdecl

\bibitem{Ye2021VisibleInfraredPR}
M.~Ye, J.~Shen, and L.~Shao, ``Visible-infrared person re-identification via
  homogeneous augmented tri-modal learning,'' \emph{IEEE Transactions on
  Information Forensics and Security}, vol.~16, pp. 728--739, 2021.

\bibitem{Wang2019RGBInfraredCP}
G.~Wang, T.~Zhang, J.~Cheng, S.~Liu, Y.~Yang, and Z.-H. Hou, ``Rgb-infrared
  cross-modality person re-identification via joint pixel and feature
  alignment,'' \emph{2019 IEEE/CVF International Conference on Computer Vision
  (ICCV)}, pp. 3622--3631, 2019.

\bibitem{Khan2016PersonRF}
F.~M. Khan and F.~Br{\'e}mond, ``Person re-identification for real-world
  surveillance systems,'' \emph{ArXiv}, vol. abs/1607.05975, 2016.

\bibitem{article}
R.~Vezzani, D.~Baltieri, and R.~Cucchiara, ``People reidentification in
  surveillance and forensics: A survey,'' \emph{ACM Computing Surveys}, 12
  2013.

\bibitem{Shen2018PersonRW}
Y.~Shen, H.~Li, S.~Yi, D.~Chen, and X.~Wang, ``Person re-identification with
  deep similarity-guided graph neural network,'' in \emph{ECCV}, 2018.

\bibitem{Xu_2018_CVPR}
J.~Xu, R.~Zhao, F.~Zhu, H.~Wang, and W.~Ouyang, ``Attention-aware compositional
  network for person re-identification,'' in \emph{Proceedings of the IEEE
  Conference on Computer Vision and Pattern Recognition (CVPR)}, June 2018.

\bibitem{Tay_2019_CVPR}
C.-P. Tay, S.~Roy, and K.-H. Yap, ``Aanet: Attribute attention network for
  person re-identifications,'' in \emph{Proceedings of the IEEE/CVF Conference
  on Computer Vision and Pattern Recognition (CVPR)}, June 2019.

\bibitem{Luo_2019_CVPR_Workshops}
H.~Luo, Y.~Gu, X.~Liao, S.~Lai, and W.~Jiang, ``Bag of tricks and a strong
  baseline for deep person re-identification,'' in \emph{Proceedings of the
  IEEE/CVF Conference on Computer Vision and Pattern Recognition (CVPR)
  Workshops}, June 2019.

\bibitem{Jin_2020_CVPR}
X.~Jin, C.~Lan, W.~Zeng, Z.~Chen, and L.~Zhang, ``Style normalization and
  restitution for generalizable person re-identification,'' in
  \emph{Proceedings of the IEEE/CVF Conference on Computer Vision and Pattern
  Recognition (CVPR)}, June 2020.

\bibitem{Ye2020CrossModalityPR}
M.~Ye, X.~Lan, and Q.~Leng, ``Cross-modality person re-identification via
  modality-aware collaborative ensemble learning,'' \emph{IEEE Transactions on
  Image Processing}, vol.~29, pp. 9387--9399, 2020.

\bibitem{Liu2016TransferringDR}
X.~Liu, L.~Song, X.~Wu, and T.~Tan, ``Transferring deep representation for
  nir-vis heterogeneous face recognition,'' \emph{2016 International Conference
  on Biometrics (ICB)}, pp. 1--8, 2016.

\bibitem{Wu2018CoupledDL}
X.~Wu, L.~Song, R.~He, and T.~Tan, ``Coupled deep learning for heterogeneous
  face recognition,'' in \emph{AAAI}, 2018.

\bibitem{Ma2019InfraredAV}
J.~Ma, Y.~Ma, and C.~Li, ``Infrared and visible image fusion methods and
  applications: A survey,'' \emph{Inf. Fusion}, vol.~45, pp. 153--178, 2019.

\bibitem{Wang2020CrossModalityPG}
G.~Wang, T.~Zhang, Y.~Yang, J.~Cheng, J.~Chang, X.~Liang, and Z.~Hou,
  ``Cross-modality paired-images generation for rgb-infrared person
  re-identification,'' in \emph{AAAI}, 2020.

\bibitem{Huang2020ProbabilityWC}
F.~Huang, L.~Zhang, Y.~Yang, and X.~Zhou, ``Probability weighted compact
  feature for domain adaptive retrieval,'' \emph{2020 IEEE/CVF Conference on
  Computer Vision and Pattern Recognition (CVPR)}, pp. 9579--9588, 2020.

\bibitem{Wu2020RGBIRPR}
A.~Wu, W.~Zheng, S.~Gong, and J.~Lai, ``Rgb-ir person re-identification by
  cross-modality similarity preservation,'' \emph{International Journal of
  Computer Vision}, vol. 128, pp. 1765--1785, 2020.

\bibitem{Ding2020MultitaskLW}
C.~Ding, K.~Wang, P.~Wang, and D.~Tao, ``Multi-task learning with coarse priors
  for robust part-aware person re-identification,'' \emph{IEEE transactions on
  pattern analysis and machine intelligence}, vol.~PP, 2020.

\bibitem{Ye2021DeepLF}
M.~Ye, J.~Shen, G.~Lin, T.~Xiang, L.~Shao, and S.~Hoi, ``Deep learning for
  person re-identification: A survey and outlook,'' \emph{IEEE transactions on
  pattern analysis and machine intelligence}, vol.~PP, 2021.

\bibitem{Wang_2019_CVPR}
Z.~Wang, Z.~Wang, Y.~Zheng, Y.-Y. Chuang, and S.~Satoh, ``Learning to reduce
  dual-level discrepancy for infrared-visible person re-identification,'' in
  \emph{Proceedings of the IEEE/CVF Conference on Computer Vision and Pattern
  Recognition (CVPR)}, June 2019.

\bibitem{Choi_2020_CVPR}
S.~Choi, S.~Lee, Y.~Kim, T.~Kim, and C.~Kim, ``Hi-cmd: Hierarchical
  cross-modality disentanglement for visible-infrared person
  re-identification,'' in \emph{Proceedings of the IEEE/CVF Conference on
  Computer Vision and Pattern Recognition (CVPR)}, June 2020.

\bibitem{Kansal2020SDLSR}
K.~Kansal, A.~V. Subramanyam, Z.~Wang, and S.~Satoh, ``Sdl:
  Spectrum-disentangled representation learning for visible-infrared person
  re-identification,'' \emph{IEEE Transactions on Circuits and Systems for
  Video Technology}, vol.~30, pp. 3422--3432, 2020.

\bibitem{Ye2018VisibleTP}
M.~Ye, Z.~Wang, X.~Lan, and P.~Yuen, ``Visible thermal person re-identification
  via dual-constrained top-ranking,'' in \emph{IJCAI}, 2018.

\bibitem{Ye2018HierarchicalDL}
M.~Ye, X.~Lan, J.~Li, and P.~Yuen, ``Hierarchical discriminative learning for
  visible thermal person re-identification,'' in \emph{AAAI}, 2018.

\bibitem{Dai2018CrossModalityPR}
P.~Dai, R.~Ji, H.~Wang, Q.~Wu, and Y.~Huang, ``Cross-modality person
  re-identification with generative adversarial training,'' in \emph{IJCAI},
  2018.

\bibitem{Hao2019HSMEHM}
Y.~Hao, N.~Wang, J.~Li, and X.~Gao, ``Hsme: Hypersphere manifold embedding for
  visible thermal person re-identification,'' in \emph{AAAI}, 2019.

\bibitem{Zhang2019AttendTT}
S.~Zhang, Y.~Yang, P.~Wang, X.~Zhang, and Y.~Zhang, ``Attend to the difference:
  Cross-modality person re-identification via contrastive correlation,''
  \emph{ArXiv}, vol. abs/1910.11656, 2019.

\bibitem{Lu_2020_CVPR}
Y.~Lu, Y.~Wu, B.~Liu, T.~Zhang, B.~Li, Q.~Chu, and N.~Yu, ``Cross-modality
  person re-identification with shared-specific feature transfer,'' in
  \emph{Proceedings of the IEEE/CVF Conference on Computer Vision and Pattern
  Recognition (CVPR)}, June 2020.

\bibitem{Ye2020BiDirectionalCT}
M.~Ye, X.~Lan, Z.~Wang, and P.~Yuen, ``Bi-directional center-constrained
  top-ranking for visible thermal person re-identification,'' \emph{IEEE
  Transactions on Information Forensics and Security}, vol.~15, pp. 407--419,
  2020.

\bibitem{Ye2020DynamicDA}
M.~Ye, J.~Shen, D.~J. Crandall, L.~Shao, and J.~Luo, ``Dynamic dual-attentive
  aggregation learning for visible-infrared person re-identification,''
  \emph{ArXiv}, vol. abs/2007.09314, 2020.

\bibitem{He2019WassersteinCL}
R.~He, X.~Wu, Z.~Sun, and T.~Tan, ``Wasserstein cnn: Learning invariant
  features for nir-vis face recognition,'' \emph{IEEE Transactions on Pattern
  Analysis and Machine Intelligence}, vol.~41, pp. 1761--1773, 2019.

\bibitem{Ye2020ImprovingNP}
M.~Ye, Y.~Cheng, X.~Lan, and H.~Zhu, ``Improving night-time pedestrian
  retrieval with distribution alignment and contextual distance,'' \emph{IEEE
  Transactions on Industrial Informatics}, vol.~16, pp. 615--624, 2020.

\bibitem{Yang2020HeterogeneousGA}
X.~Yang, C.~Deng, T.~Liu, and D.~Tao, ``Heterogeneous graph attention network
  for unsupervised multiple-target domain adaptation.'' \emph{IEEE transactions
  on pattern analysis and machine intelligence}, vol.~PP, 2020.

\bibitem{Kang2019ContrastiveAN}
G.~Kang, L.~Jiang, Y.~Yang, and A.~Hauptmann, ``Contrastive adaptation network
  for unsupervised domain adaptation,'' \emph{2019 IEEE/CVF Conference on
  Computer Vision and Pattern Recognition (CVPR)}, pp. 4888--4897, 2019.

\bibitem{Wang2020DeepMM}
P.~Wang, Z.~Zhao, F.~Su, Y.~Zhao, H.~Wang, L.~Yang, and Y.~Li, ``Deep
  multi-patch matching network for visible thermal person re-identification,''
  \emph{IEEE Transactions on Multimedia}, pp. 1--1, 2020.

\bibitem{Xu_2020_CVPR}
R.~Xu, P.~Liu, L.~Wang, C.~Chen, and J.~Wang, ``Reliable weighted optimal
  transport for unsupervised domain adaptation,'' in \emph{Proceedings of the
  IEEE/CVF Conference on Computer Vision and Pattern Recognition (CVPR)}, June
  2020.

\bibitem{Wei2021FlexibleBP}
Z.~Wei, X.~Yang, N.~Wang, and X.~Gao, ``Flexible body partition-based
  adversarial learning for visible infrared person re-identification.''
  \emph{IEEE transactions on neural networks and learning systems}, vol.~PP,
  2021.

\bibitem{Zhang2021MultiScaleCN}
C.~Zhang, H.-C. Liu, W.~Guo, and M.~Ye, ``Multi-scale cascading network with
  compact feature learning for rgb-infrared person re-identification,''
  \emph{2020 25th International Conference on Pattern Recognition (ICPR)}, pp.
  8679--8686, 2021.

\bibitem{Courty2017OptimalTF}
N.~Courty, R.~Flamary, D.~Tuia, and A.~Rakotomamonjy, ``Optimal transport for
  domain adaptation,'' \emph{IEEE Transactions on Pattern Analysis and Machine
  Intelligence}, vol.~39, pp. 1853--1865, 2017.

\bibitem{Wu_2017_ICCV}
A.~Wu, W.-S. Zheng, H.-X. Yu, S.~Gong, and J.~Lai, ``Rgb-infrared
  cross-modality person re-identification,'' in \emph{Proceedings of the IEEE
  International Conference on Computer Vision (ICCV)}, Oct 2017.

\bibitem{kang2019person}
J.~K. Kang, T.~M. Hoang, and K.~R. Park, ``Person re-identification between
  visible and thermal camera images based on deep residual cnn using single
  input,'' \emph{IEEE Access}, vol.~7, pp. 57\,972--57\,984, 2019.

\bibitem{Tekeli_2019_ICCV}
N.~Tekeli and A.~Burak~Can, ``Distance based training for cross-modality person
  re-identification,'' in \emph{Proceedings of the IEEE/CVF International
  Conference on Computer Vision (ICCV) Workshops}, Oct 2019.

\bibitem{feng2019learning}
Z.~Feng, J.~Lai, and X.~Xie, ``Learning modality-specific representations for
  visible-infrared person re-identification,'' \emph{IEEE Transactions on Image
  Processing}, vol.~29, pp. 579--590, 2019.

\bibitem{Ye2019ModalityawareCL}
M.~Ye, X.~Lan, and Q.~Leng, ``Modality-aware collaborative learning for visible
  thermal person re-identification,'' \emph{Proceedings of the 27th ACM
  International Conference on Multimedia}, 2019.

\bibitem{Zhu2020HeteroCenterLF}
Y.~Zhu, Z.~Yang, L.~Wang, S.~Zhao, X.~Hu, and D.~Tao, ``Hetero-center loss for
  cross-modality person re-identification,'' \emph{ArXiv}, vol. abs/1910.09830,
  2020.

\bibitem{Liu2020ParametersSE}
H.~Liu and X.~Tan, ``Parameters sharing exploration and hetero-center based
  triplet loss for visible-thermal person re-identification,'' \emph{ArXiv},
  vol. abs/2008.06223, 2020.

\bibitem{Varior2016ASL}
R.~Varior, B.~Shuai, J.~Lu, D.~Xu, and G.~Wang, ``A siamese long short-term
  memory architecture for human re-identification,'' \emph{ArXiv}, vol.
  abs/1607.08381, 2016.

\bibitem{Sun2018BeyondPM}
Y.~Sun, L.~Zheng, Y.~Yang, Q.~Tian, and S.~Wang, ``Beyond part models: Person
  retrieval with refined part pooling,'' in \emph{ECCV}, 2018.

\bibitem{Li_2017_CVPR}
D.~Li, X.~Chen, Z.~Zhang, and K.~Huang, ``Learning deep context-aware features
  over body and latent parts for person re-identification,'' in
  \emph{Proceedings of the IEEE Conference on Computer Vision and Pattern
  Recognition (CVPR)}, July 2017.

\bibitem{BAI2020107036}
\BIBentryALTinterwordspacing
X.~Bai, M.~Yang, T.~Huang, Z.~Dou, R.~Yu, and Y.~Xu, ``Deep-person: Learning
  discriminative deep features for person re-identification,'' \emph{Pattern
  Recognition}, vol.~98, p. 107036, 2020. [Online]. Available:
  \url{http://www.sciencedirect.com/science/article/pii/S0031320319303395}
\BIBentrySTDinterwordspacing

\bibitem{Insafutdinov2016DeeperCutAD}
E.~Insafutdinov, L.~Pishchulin, B.~Andres, M.~Andriluka, and B.~Schiele,
  ``Deepercut: A deeper, stronger, and faster multi-person pose estimation
  model,'' in \emph{ECCV}, 2016.

\bibitem{Sun_2019_CVPR}
K.~Sun, B.~Xiao, D.~Liu, and J.~Wang, ``Deep high-resolution representation
  learning for human pose estimation,'' in \emph{Proceedings of the IEEE/CVF
  Conference on Computer Vision and Pattern Recognition (CVPR)}, June 2019.

\bibitem{kantorovich2006problem}
L.~V. Kantorovich, ``On a problem of monge,'' \emph{J. Math. Sci.(NY)}, vol.
  133, p. 1383, 2006.

\bibitem{Liu2021OptimalTD}
Z.-H. Liu, L.-B. Jiang, H.-L. Wei, L.~Chen, and X.-H. Li, ``Optimal
  transport-based deep domain adaptation approach for fault diagnosis of
  rotating machine,'' \emph{IEEE Transactions on Instrumentation and
  Measurement}, vol.~70, pp. 1--12, 2021.

\bibitem{Chen2020GraphOT}
L.~Chen, Z.~Gan, Y.~Cheng, L.~Li, L.~Carin, and J.~jing Liu, ``Graph optimal
  transport for cross-domain alignment,'' \emph{ArXiv}, vol. abs/2006.14744,
  2020.

\bibitem{Yuan2020WeaklySC}
S.~Yuan, K.~Bai, L.~Chen, Y.~zhe Zhang, C.~Tao, C.~Li, G.~Wang, R.~Henao, and
  L.~Carin, ``Weakly supervised cross-domain alignment with optimal
  transport,'' \emph{ArXiv}, vol. abs/2008.06597, 2020.

\bibitem{Chen2019UNITERLU}
Y.-C. Chen, L.~Li, L.~Yu, A.~E. Kholy, F.~Ahmed, Z.~Gan, Y.~Cheng, and J.~Liu,
  ``Uniter: Learning universal image-text representations,'' \emph{ArXiv}, vol.
  abs/1909.11740, 2019.

\bibitem{Li_2020_CVPR}
M.~Li, Y.-M. Zhai, Y.-W. Luo, P.-F. Ge, and C.-X. Ren, ``Enhanced transport
  distance for unsupervised domain adaptation,'' in \emph{Proceedings of the
  IEEE/CVF Conference on Computer Vision and Pattern Recognition (CVPR)}, June
  2020.

\bibitem{seguy2017large}
V.~Seguy, B.~B. Damodaran, R.~Flamary, N.~Courty, A.~Rolet, and M.~Blondel,
  ``Large-scale optimal transport and mapping estimation,'' \emph{arXiv
  preprint arXiv:1711.02283}, 2017.

\bibitem{Damodaran_2018_ECCV}
B.~B. Damodaran, B.~Kellenberger, R.~Flamary, D.~Tuia, and N.~Courty,
  ``Deepjdot: Deep joint distribution optimal transport for unsupervised domain
  adaptation,'' in \emph{Proceedings of the European Conference on Computer
  Vision (ECCV)}, September 2018.

\bibitem{Lee2019HierarchicalOT}
J.~Lee, M.~Dabagia, E.~L. Dyer, and C.~Rozell, ``Hierarchical optimal transport
  for multimodal distribution alignment,'' in \emph{NeurIPS}, 2019.

\bibitem{Velickovic2018GraphAN}
P.~Velickovic, G.~Cucurull, A.~Casanova, A.~Romero, P.~Li{\`o}, and Y.~Bengio,
  ``Graph attention networks,'' \emph{ArXiv}, vol. abs/1710.10903, 2018.

\bibitem{kipf2016semi}
T.~N. Kipf and M.~Welling, ``Semi-supervised classification with graph
  convolutional networks,'' \emph{arXiv preprint arXiv:1609.02907}, 2016.

\bibitem{huang2019signed}
J.~Huang, H.~Shen, L.~Hou, and X.~Cheng, ``Signed graph attention networks,''
  in \emph{International Conference on Artificial Neural Networks}.\hskip 1em
  plus 0.5em minus 0.4em\relax Springer, 2019, pp. 566--577.

\bibitem{Lee2018AttentionMI}
J.~B. Lee, R.~A. Rossi, S.~Kim, N.~K. Ahmed, and E.~Koh, ``Attention models in
  graphs: A survey,'' \emph{arXiv: Artificial Intelligence}, 2018.

\bibitem{Zhang_2021_CVPR}
Z.~Zhang, H.~Zhang, and S.~Liu, ``Person re-identification using heterogeneous
  local graph attention networks,'' in \emph{Proceedings of the IEEE/CVF
  Conference on Computer Vision and Pattern Recognition (CVPR)}, June 2021, pp.
  12\,136--12\,145.

\bibitem{Zhu2020ASG}
Y.~Zhu, Z.~Zha, T.~Zhang, J.~Liu, and J.~Luo, ``A structured graph attention
  network for vehicle re-identification,'' \emph{Proceedings of the 28th ACM
  International Conference on Multimedia}, 2020.

\bibitem{bao2019masked}
L.~Bao, B.~Ma, H.~Chang, and X.~Chen, ``Masked graph attention network for
  person re-identification,'' in \emph{Proceedings of the IEEE/CVF Conference
  on Computer Vision and Pattern Recognition Workshops}, 2019, pp. 0--0.

\bibitem{zhang2021person}
Z.~Zhang, H.~Zhang, and S.~Liu, ``Person re-identification using heterogeneous
  local graph attention networks,'' in \emph{Proceedings of the IEEE/CVF
  Conference on Computer Vision and Pattern Recognition}, 2021, pp.
  12\,136--12\,145.

\bibitem{Ye2020AugmentationIA}
M.~Ye, J.~Shen, X.~Zhang, P.~Yuen, and S.-F. Chang, ``Augmentation invariant
  and instance spreading feature for softmax embedding.'' \emph{IEEE
  transactions on pattern analysis and machine intelligence}, vol.~PP, 2020.

\bibitem{abdellali2019robust}
H.~Abdellali, R.~Frohlich, and Z.~Kato, ``Robust absolute and relative pose
  estimation of a central camera system from 2d-3d line correspondences,'' in
  \emph{Proceedings of the IEEE/CVF International Conference on Computer Vision
  Workshops}, 2019, pp. 0--0.

\bibitem{Li2020XModality}
D.~Li, X.~Wei, X.~Hong, and Y.~Gong, ``Infrared-visible cross-modal person
  re-identification with an x modality,'' in \emph{AAAI}, 2020.

\bibitem{Shen2018WassersteinDG}
J.~Shen, Y.~Qu, W.~Zhang, and Y.~Yu, ``Wasserstein distance guided
  representation learning for domain adaptation,'' in \emph{AAAI}, 2018.

\bibitem{Zanfir2018DeepLO}
A.~Zanfir and C.~Sminchisescu, ``Deep learning of graph matching,'' \emph{2018
  IEEE/CVF Conference on Computer Vision and Pattern Recognition}, pp.
  2684--2693, 2018.

\bibitem{Arjovsky2017WassersteinGA}
M.~Arjovsky, S.~Chintala, and L.~Bottou, ``Wasserstein generative adversarial
  networks,'' in \emph{ICML}, 2017.

\bibitem{Peyr2019ComputationalOT}
G.~Peyr{\'e} and M.~Cuturi, ``Computational optimal transport,'' \emph{Found.
  Trends Mach. Learn.}, vol.~11, pp. 355--607, 2019.

\bibitem{Peyr2016GromovWassersteinAO}
G.~Peyr{\'e}, M.~Cuturi, and J.~Solomon, ``Gromov-wasserstein averaging of
  kernel and distance matrices,'' in \emph{ICML}, 2016.

\bibitem{Hadsell2006DimensionalityRB}
R.~Hadsell, S.~Chopra, and Y.~LeCun, ``Dimensionality reduction by learning an
  invariant mapping,'' \emph{2006 IEEE Computer Society Conference on Computer
  Vision and Pattern Recognition (CVPR'06)}, vol.~2, pp. 1735--1742, 2006.

\bibitem{Xu2019ScalableGL}
H.~Xu, D.~Luo, and L.~Carin, ``Scalable gromov-wasserstein learning for graph
  partitioning and matching,'' \emph{ArXiv}, vol. abs/1905.07645, 2019.

\bibitem{Yang_2019_CVPR}
W.~Yang, H.~Huang, Z.~Zhang, X.~Chen, K.~Huang, and S.~Zhang, ``Towards rich
  feature discovery with class activation maps augmentation for person
  re-identification,'' in \emph{Proceedings of the IEEE/CVF Conference on
  Computer Vision and Pattern Recognition (CVPR)}, June 2019.

\bibitem{Vaswani2017AttentionIA}
A.~Vaswani, N.~Shazeer, N.~Parmar, J.~Uszkoreit, L.~Jones, A.~N. Gomez,
  L.~Kaiser, and I.~Polosukhin, ``Attention is all you need,'' in \emph{NIPS},
  2017.

\bibitem{Wang2020DeepMF}
Y.~Wang, W.~Huang, F.-C. Sun, T.~Xu, Y.~Rong, and J.~Huang, ``Deep multimodal
  fusion by channel exchanging,'' \emph{ArXiv}, vol. abs/2011.05005, 2020.

\bibitem{dalal2005histograms}
N.~Dalal and B.~Triggs, ``Histograms of oriented gradients for human
  detection,'' in \emph{2005 IEEE computer society conference on computer
  vision and pattern recognition (CVPR'05)}, vol.~1.\hskip 1em plus 0.5em minus
  0.4em\relax IEEE, 2005, pp. 886--893.

\bibitem{liao2015person}
S.~Liao, Y.~Hu, X.~Zhu, and S.~Z. Li, ``Person re-identification by local
  maximal occurrence representation and metric learning,'' in \emph{Proceedings
  of the IEEE conference on computer vision and pattern recognition}, 2015, pp.
  2197--2206.

\bibitem{Nguyen2017PersonRS}
T.~D. Nguyen, H.~Hong, K.~Kim, and K.~Park, ``Person recognition system based
  on a combination of body images from visible light and thermal cameras,''
  \emph{Sensors (Basel, Switzerland)}, vol.~17, 2017.

\bibitem{He2016DeepRL}
K.~He, X.~Zhang, S.~Ren, and J.~Sun, ``Deep residual learning for image
  recognition,'' \emph{2016 IEEE Conference on Computer Vision and Pattern
  Recognition (CVPR)}, pp. 770--778, 2016.

\bibitem{Deng2009ImageNetAL}
J.~Deng, W.~Dong, R.~Socher, L.~Li, K.~Li, and L.~Fei-Fei, ``Imagenet: A
  large-scale hierarchical image database,'' in \emph{CVPR}, 2009.

\bibitem{Lin2014MicrosoftCC}
T.-Y. Lin, M.~Maire, S.~Belongie, J.~Hays, P.~Perona, D.~Ramanan,
  P.~Doll{\'a}r, and C.~L. Zitnick, ``Microsoft coco: Common objects in
  context,'' \emph{ArXiv}, vol. abs/1405.0312, 2014.

\bibitem{Wang2020HighOrderIM}
G.~Wang, S.~Yang, H.~Liu, Z.~Wang, Y.~Yang, S.~Wang, G.~Yu, E.~Zhou, and
  J.~Sun, ``High-order information matters: Learning relation and topology for
  occluded person re-identification,'' \emph{2020 IEEE/CVF Conference on
  Computer Vision and Pattern Recognition (CVPR)}, pp. 6448--6457, 2020.

\bibitem{Maaten2008VisualizingDU}
L.~V.~D. Maaten and G.~E. Hinton, ``Visualizing data using t-sne,''
  \emph{Journal of Machine Learning Research}, vol.~9, pp. 2579--2605, 2008.

\end{thebibliography}

%\end{thebibliography}


%\end{IEEEbiography}


%\newpage



%\vfill


\end{document}